\documentclass{article}

\PassOptionsToPackage{numbers,compress}{natbib}

    \usepackage[final]{neurips_2021}

\usepackage{microtype}
\usepackage{graphicx}
\usepackage{subfigure}
\usepackage{booktabs} 
\usepackage{natbib}

\usepackage[utf8]{inputenc} 
\usepackage[hypertexnames=false]{hyperref}      
\usepackage[dvipsnames]{xcolor}

\hypersetup{
    colorlinks,
    linkcolor={red!50!black},
    citecolor={blue!50!black},
    urlcolor={blue!80!black}
}

\renewcommand{\paragraph}[1]{\vspace{-0.1em}\noindent \textbf{#1}}

\usepackage{inconsolata}
\usepackage{url}            
\usepackage{amsfonts}       
\usepackage{nicefrac}       
\usepackage{marginalia}
\usepackage{xr-hyper}
\usepackage{amsthm}
\usepackage{array,amssymb,latexsym,amssymb,epsfig,epsf,amsmath,cite,accents,float,mathrsfs,verbatim,bm,color,enumitem,bbm}
\graphicspath{{figures/}}
\usepackage[algo2e,linesnumbered,noend]{algorithm2e}
\SetKwInput{KwEn}{Encoding}
\SetKwInput{KwDe}{Decoding}
\usepackage{wrapfig}
\usepackage{transparent}

\usepackage[capitalise]{cleveref}
\usepackage[font=small,labelfont=bf]{caption}
\usepackage{mathtools}
\usepackage{adjustbox}

\crefname{ineq}{}{}
\creflabelformat{ineq}{#2{\upshape(#1)}#3} 
\crefname{prob}{}{}
\creflabelformat{prob}{#2{\upshape(#1)}#3} 

\newtheorem{lemma}{Lemma}

\newtheorem{theorem}{Theorem}
\newtheorem{corollary}{Corollary}
\newtheorem{remark}{Remark}
\newtheorem{definition}{Definition}

\newtheorem{proposition}{Proposition}
\newtheorem{assumption}{Assumption}

\newcolumntype{C}[1]{>{\centering\arraybackslash}m{#1}}

\def\R{\mathbb{R}}

\def\integers{\mathbb{Z}}

\def\abf{{\bf a}}
\def\bbf{{\bf b}}

\def\sbf{{\bf s}}

\def\ubf{{\bf u}}
\def\vbf{{\bf v}}
\def\wbf{{\bf w}}
\def\xbf{{\bf x}}

\def\zbf{{\bf z}}

\def\xbf{{\bf x}}

\def\Cc{{\cal C}}

\def\Ec{{\cal E}}

\def\Hc{{\cal H}}

\def\Nc{{\cal N}}

\def\nn{\nonumber}
\def\beq{\begin{equation}}
\def\eeq{\end{equation}}
\def\beqa{\begin{eqnarray}}
\def\eeqa{\end{eqnarray}}
\def\balign{\begin{align}}
\def\ealign{\end{align}}
\def\bpr{\begin{proof}}
\def\epr{\end{proof}}
\def\bth{\begin{theorem}}
\def\eth{\end{theorem}}
\def\blm{\begin{lemma}}
\def\elm{\end{lemma}}
\def\bprop{\begin{proposition}}
\def\eprop{\end{proposition}}
\def\bcr{\begin{corollary}}
\def\ecr{\end{corollary}}
\def\ie{{\it i.e.,\ \/}}
\def\eg{{\it e.g.,\ \/}}

\def\tr{{\rm tr}}
\def\vector{{\rm vec}}

\def\E{\mathbb{E}}

\def\and {{\rm and}}

\newcommand{\op}{{overparameterization}\xspace}
\newcommand{\Op}{{Overparameterization}\xspace}
\newcommand{\oped}{{overparameterized}\xspace}

\def \l{\left}
\def \r{\right}
\def \a{\alpha}
\def \b{\beta}
\def \g{\gamma}
\def \z{\zeta}
\def\o{\omega}
\def \t{\tau}
\def \s{\sigma}

\def \R{\mathbb{R}}

\def \ol{\overline}
\def \tr{\operatorname{trace}}

\def \der{\operatorname{d}\hspace{-1pt}} 
\def \Der{\nabla\hspace{-1pt}}

\def \ball{\operatorname{ball}}
\def\m{\max}

\def\ra{\rightarrow}
\def\da{\downarrow}
\def\bl{\left\{}
\def\br{\right\}}

\allowdisplaybreaks
\graphicspath{{figs/}}

\usepackage{xcolor}
\usepackage[showdeletions]{color-edits}		
\addauthor[Thomas]{TA}{Purple!80}

\newcommand*\samethanks[1][\value{footnote}]{\footnotemark[#1]}
\title{Subquadratic \Op for\\ Shallow Neural Networks}

\author{%
Chaehwan Song$^1$\thanks{Equal contributions.}
    \And
    Ali Ramezani-Kebrya$^1$\samethanks{}
    \AND 
    Thomas Pethick$^1$
    \And
    Armin Eftekhari$^2$\thanks{This work was done while Armin Eftekhari was at EPFL.}
    \And
    Volkan Cevher$^1$ 
       \And
    \vspace*{-.7cm}
    \\
$^1$Laboratory for Information and Inference
Systems (LIONS), EPFL
 \quad
    $^2$Umea University\\[.3cm]
    \texttt{ali.ramezani@epfl.ch}
     
}

\begin{document}

\maketitle

\begin{abstract}
\Op refers to the important phenomenon where the width of a neural network is chosen such that learning algorithms can provably attain zero loss in nonconvex training. The existing theory establishes such global convergence using various initialization strategies, training modifications, and width scalings. In particular, the state-of-the-art results require the width to scale quadratically with the number of training data under standard initialization strategies used in practice for best generalization performance. In contrast, the most recent results obtain linear scaling either with requiring initializations that lead to the ``lazy-training'',  or training only a single layer. In this work, we provide an analytical framework that allows us to adopt standard initialization strategies, possibly avoid lazy training, and train all layers simultaneously in basic shallow neural networks while attaining  a desirable subquadratic scaling on the network width. We achieve the desiderata via Polyak-{\L}ojasiewicz condition, smoothness, and standard assumptions on data, and use tools from random matrix theory.

\end{abstract}

\section{Introduction}\label{sec:intro}
{Training a neural network involves solving a nonconvex optimization problem, which, in theory, might trap first-order methods such as gradient descent to fall in bad local minima or saddle points. However, empirical evidence suggests that first-order methods with random initialization can consistently find a global minimum, even with randomized labels~\citep{zhang2017understanding}. Demystifying this observation is of central interest to deep learning.}

Recently, a line of research \citep{zou2018stochastic, brutzkus2017globally, du2018power, li2018learning, song2019quadratic, du2018gradient, oymak2019towards} suggests that such an empirical success can possibly be explained by the \emph{\op} of neural networks, whose number of parameters  exceeds the number of training data $n$. In particular, gradient descent converges linearly fast to a global optimum  in a number of problems with models that have wide hidden layers \citep{zou2018stochastic,du2018gradient,song2019quadratic}.

{Despite of these remarkable results, the natural key question \emph{``How much should we overparameterize a neural network?''} remains open even for the toy example of two-layer neural networks. On one hand, it is widely accepted that, for two-layer neural networks, the number of parameters should grow linearly with $n$ (\eg \citep{kawaguchi2019gradient,oymak2019towards}). On the other hand, theoretical results either require much more parameters, or they are established under restrictive settings. Specifically,
\begin{itemize}[leftmargin=*,itemsep=0ex]
\item \citet{kawaguchi2019gradient} has proven the ideal $\tilde{\Omega}(n)$ scaling for deep neural networks. However,  they apply gradient descent only to the last layer, which is not the case in practical scenarios.
\item A similar issue exists in \citep{song2019quadratic, oymak2019towards}, where the authors have shown that $\tilde{\Omega}(n^2)$ parameters suffice for two-layer neural networks, but only the first layers are trained. Furthermore,  even with infinite width, \citet{oymak2019towards} cannot guarantee zero training error with probability approaching to one. 
\end{itemize}

The goal of this paper is to close the gap between  theory and  practice, without resorting to unrealistic assumptions such as those discussed above. We sharpen the results of \citet{oymak2019towards} by proving that, with proper random initialization of each layer, training error approaches to zero with high probability, exponentially fast in the width of the network. In addition, we show that only $\tilde{\Omega}(n^{\frac{3}{2}})$ parameters suffice such that gradient descent converges to a global minimum with linear rate,  which improves upon the state-of-the-art by a factor of $\tilde{O}(n^\frac{1}{2})$. We summarize the bounds on the number of parameters in terms of $n$ in Table~\ref{tab:acc}.

While our analysis on gradient descent focuses on training error, it has been observed that \op can lead to poor \emph{generalization}. In particular,  \citep{chizat2019lazy, yehudai2019power, ghorbani2019limitations} have observed the phenomenon of \emph{lazy training}. \citet{chizat2019lazy} has explained lazy training as a model behaves similar to its linearization around the initialization. It is known that an \oped neural network is likely to be trapped in the lazy regime since the parameters will hardly vary over the course of training with gradient descent ~\citep{du2018gradient, li2018learning, zou2018stochastic}. The same phenomenon has been observed for infinitely wide neural networks~\citep{jacot2018neural}. In this paper, we provide theoretical guidance to possibly avoid lazy training through proper initialization.  Experimental results confirm that lazy training might be avoided with our theoretically inspired initialization so that the issues reported in~\citep{chizat2019lazy} do not apply. 

\subsection{Summary of contributions}
\begin{itemize}[leftmargin=*,itemsep=0ex]

\item We first focus on a general minimization problem assuming that the loss function satisfies Polyak-{\L}ojasiewicz (PL) condition. We find sufficient conditions in terms of initialization for the convergence of gradient flow and gradient descent to a global minimum.
    \item We then focus on the special problem of training a two-layer neural network with quadratic loss and smooth activation, and show that $\tilde{\Omega}(n^{\frac{3}{2}})$ parameters are sufficient for gradient descent to converge to a global minimum with linear rate and probability approaching to one.  We achieve \textit{linear scaling} for the width when the number of input features is in $\tilde\Omega(\sqrt{n}) $.  
    \item We theoretically guide how to initialize the parameters of a neural network in the \oped regime of interest while possibly avoiding lazy training.     
\end{itemize} 


\subsection{Further related work}  
\begin{table}
\centering
 \caption{Scaling with the number of training data in the \op regime. QL=quadratic loss, CLL=convex and Lipschitz loss,  SD=separable data.}
 \begin{adjustbox}{center}
	\begin{tabular}{c c c c c c} \label{tab:acc}
	\textbf{Depth}& \textbf{Algorithm}& \textbf{Setting}& \textbf{Activation}& \textbf{Scaling}&\textbf{Reference}\\
	\hline
	2&GD on layer 1&QL&ReLU& $\tilde{\Omega}(n^2)$&\citet{oymak2019towards}\\
	\hline
	$L$&GD on layer $L$&CLL&ReLU&$\tilde{\Omega}(n)$& \citet{kawaguchi2019gradient}\\
	\hline
	2&GD&SD&ReLU&$\tilde{\Omega}(n^2)$&\citet{song2019quadratic}\\
	\hline
	2&GD&SD and QL&ReLU&$\tilde{\Omega}(n^6)$&\citet{du2018gradient}\\
	\hline
	$L$&GD&SD and QL&ReLU&$\Omega(n^{8}L^{12})$&\citet{zou2019improved}\\
	\hline
	2&GD&QL &Smooth&$\tilde\Omega(n^{\frac{3}{2}})$&\textbf{This paper}\\
	\end{tabular}
 \end{adjustbox}
\end{table}

In terms of techniques, our paper is closely related to \citep{ oymak2018overparameterized,oymak2019towards}. Similar to our Theorem~\ref{thm:shallow-net-ex2}, \citet[Theorem 2.1]{oymak2019towards}  showed that gradient descent converges with linear rate when the Jacobian of the nonlinear mapping has smooth deviations, and the number of parameters grows quadratically with  $n$.  However, \citet{oymak2019towards} assumed that gradient descent  updates only the first layer. In this paper, we consider the case where gradient descent  updates both layers simultaneously, and show that it suffices to have $\tilde{\Omega}(n^{\frac{3}{2}})$ parameters with a linear rate of convergence. 

ReLU is an important instance of activation functions that does not satisfy the smoothness assumption. A line of research aims to relax this assumption by instead assuming the data is separable.
For shallow neural networks, \citet{du2018gradient} proved that gradient descent finds a global minimum if the width of the network scales $\tilde{\Omega}(n^6)$ assuming that no two data points are parallel. In a similar setting,  \citet{song2019quadratic} established convergence to a global minimum with the sufficient width of $\tilde{\Omega}(n^2)$.
As a result, in the absence of the smoothness assumption, these papers 
require substantially more number of parameters to guarantee convergence to a global minimum.

The theoretical bounds for deep neural networks are even worse. For instance,~\citet{allen2018convergence} required the total number of parameters of $ \Omega(n^{24}L^{12})$  where $L$ is the number of layers. \citet{zou2019improved} improved the scaling to $\Omega(n^{8}L^{12})$. In our setting, \ie $L=2$, these bounds become vacuous in most interesting regimes. Further, in~\citep{kawaguchi2019gradient},  the authors showed that $\tilde{\Omega}(n)$ parameters is enough to achieve global convergence under the assumption that gradient descent updates only the last layer, which essentially reduces the problem to a simple least-squares regression.
 
Recently, \citet{ji2019polylogarithmic, chen2019much} showed that a polylogarithmic width suffices to achieve convergence for shallow and deep neural networks in an ergodic sense. We note that this is a weaker notion of convergence compared to the one we consider.

\citet{oymak2020noise} showed that gradient descent along with early stopping are robust to label noise on a constant fraction of labels in an \oped network. However, only the first layer is optimized in \citep{oymak2020noise}.  For possibly \oped and linear networks, \citet{Eftekhari2020linear} showed that gradient flow can successfully avoid lazy training assuming that the network has a layer with a single neuron. We note that our analysis does not require those restrictions.

Under an assumption similar to PL condition, \citet{zou2018stochastic} studied the problem of binary classification for a deep network with ReLU activation, which is a different problem compared to ours. In~\citep{Su2019approx}, the authors proved that gradient descent with \op achieves zero-approximation when the underlying function that generates the labels has low-rank approximation. Their scaling requires perfect information about the target function, which is not the case in our paper.  Under a variant of Xavier initialization,~\citet{Daniely2020}  found near optimal scaling for a binary classification problem trained by stochastic gradient descent. We note that the setting considered in our paper is more challenging than binary classification. Our results establish a new state-of-the-art on the required number of parameters in a nonrestrictive setting when both layers are trained at the same time.  Recently, \citet{Pyr2020} obtained subquadratic scaling for a deep neural network with pyramidal structure under an initialization that leads to lazy training. Our results do not have such restrictions.  

Mean-field analysis was used to approximate a target distribution of parameters of  a neural network by the empirical distributions~\citep{MFT2019,MFT2020}. However, these results do not provide useful bounds on the scaling in terms of $n$, which is our focus in this paper.

\citet{liu2020loss} established global convergence when the function to minimize  satisfies a variant of PL condition (local PL condition) assuming the map is Lipschitz continuous, which is not the case in our paper. \citet{liu2020linearity} characterized the constancy of the neural tangent kernel via scaling properties of the norm of the Hessian matrix of the network. In this work, we focus on obtaining a sufficient number of parameters for gradient descent to converge to a global minimum with linear rate.

\paragraph{Notation.} We use  $\|\cdot\|$ to represent the Euclidean norm of a vector and Frobenius norm of a matrix. We use $\Der$ to denote the Jacobian of a vector-valued and gradient of a scalar-valued function and $\Der \Phi(a)\bl b\br$ to represent the directional derivative of $\Phi$ along $b$.  We use $\odot$ and $\otimes$ to denote the Hadamard (entry-wise) product and  Kronecker product, respectively.  For $A\in \R^{m\times n}$ and $t\in\integers_{+}$,   we denote $A^{*t} \in \R^{m^t \times n}$ with its $a$-th column defined as $\vector(x_a \otimes \cdots \otimes x_a) \in \R^{m^t}$. We use lower-case bold font to denote vectors. Sets and scalars are represented by calligraphic and standard fonts, respectively. We use $[n]$ to denote $\{1,\cdots,n\}$ for an integer $n$.  We use $\tilde{O}$ and $\tilde\Omega$ to hide logarithmic factors and use $\lesssim$ to ignore terms up to constant and logarithmic factors. 

\section{Problem, definitions, and assumptions \label{sec:setup}}
In this section, we set up a general compositional optimization problem.  Then we focus on the special case of shallow neural networks in Section~\ref{sec:nn}. 

Let $\wbf\in\R^d$ denote a parameter vector where $d$ denotes the number of parameters.  In a neural network,  $\wbf$ consists of weights and biases of all layers. We consider the minimization problem
\begin{align}
\min_{\wbf\in \R^{d}} h(\wbf)
\label{eq:main}
\end{align}
where $h:\R^{d} \rightarrow\R_+$ is the composition of a  loss function $f:\R^{\tilde d} \rightarrow\R_+$ and a nonlinear and nonconvex function $\Phi:\R^{d} \rightarrow\R^{\tilde d}$:  
\begin{align}
h(\wbf) = f(\Phi(\wbf)) = f(\zbf)
\label{eq:defn-h}
\end{align} where $\zbf = \Phi(\wbf)$.

Before providing the details, let us highlight the simple idea behind the argument (see also~\citep{oymak2018overparameterized}). Let $\wbf_0$ and $\ol \wbf$ denote the initial point and limit point when the gradient descent algorithm is run with some learning rate, respectively.  The precise formulation of gradient descent is provided in Section~\ref{sec:gd}. Let $\Der \Phi^* (\ol \wbf):\R^{\tilde d}\rightarrow\R^{d}$ denote the adjoint operator of $\Der \Phi(\ol \wbf)$.  Since $\ol \wbf$ is a first-order stationary point of $h$, we have 
\begin{align}
0 & = \Der h(\ol \wbf)=  \Der \Phi^*(\ol \wbf) \bl  \Der f(\ol{\zbf}) \br\nn 
\end{align}
where $\ol{\zbf} = \Phi(\ol \wbf)$. Suppose that  $\Der \Phi^*(\ol \wbf)$ is a nonsingular operator.Then  $\Der f(\ol{\zbf}) = 0$. If $\ol \zbf$ is a global minimizer of $f$, then  $\ol \wbf$ is a global minimizer of $h$.  To prove global convergence, it suffices to show that $\Der\Phi^*$ is nonsingular within a neighborhood of the initialization $\wbf_0$, and that points reached by gradient descent remain  within this neighborhood. We will prove that both statements hold with high probability for shallow neural networks.

We first define two notions that are useful to state a key lemma for our main results:
\begin{definition}[Near-isometry]\label{cnd:iso-jac}
A	linear mapping $T : \R^{d_1} \rightarrow \R^{d_2}$ is $(\mu,\nu)$-near-isometry if there exist $0< \mu \le \nu$ such that  
	\begin{align}\label{eq:isometry-of-DPhi}
	\mu  \le \s_{\min}(T) \le  \s_{\max}(T) \le \nu.
	\end{align}
\end{definition}

\begin{definition}[Smoothness]  Let $\b_{\psi}>0$. A function $\psi:  \R^{d_1} \rightarrow \R^{d_2}$ is $\b_{\psi}$-smooth,  if for all $\ubf, \vbf \in \R^{d_1}$, we have 
\begin{align}\label{eq:lips-cnt-DPhi}
\s_{\max}( \Der \psi(\ubf) - \Der \psi(\vbf) ) \leq \beta_{\psi} \|\ubf - \vbf \|.
\end{align}
\end{definition}

The following lemma shows that a smooth function, which is near-isometry at initialization, remains near-isometry for all nearby points of the initialization. 

\blm \label{lem:Phi-singular-bound} Suppose that $\Phi$ is $\b_{\Phi}$-smooth and $\Der \Phi^*(\wbf_0)$ is $(\mu_\Phi,\nu_\Phi)$-near-isometry. Then, for all $\wbf\in \ball(\wbf_0,\rho_\Phi)$, we have 
	\begin{align}\label{eq:iso}
	\frac{\mu_\Phi}{2} \leq \s_{\min}(\Der \Phi^*(\wbf))\le \s_{\max}(\Der \Phi^*(\wbf)) \leq \frac{3\nu_\Phi}{2}
	\end{align} where 
	\begin{align}\label{eq:perturb-assump} 
	\rho_\Phi = \frac{\mu_\Phi}{2\beta_\Phi }. 
	\end{align}
\elm

Intuitively, if $\Der \Phi^*(\wbf_0)$ is a $(\mu_\Phi,\nu_\Phi)$-near-isometry, then one would  expect $\Der \Phi^*$ to remain near-isometry for all nearby points.  

\begin{definition}[PL condition~\citep{bolte2017error}]
A function $\psi:  \R^{d_1} \rightarrow \R$ satisfies the PL condition if there exists $\a_{\psi} >0$ such that, for all  $\ubf \in \R^{d_1}$, we have 
\begin{align}\label{eq:pl}
\psi(\ubf) & \le  \frac{\| \Der \psi(\ubf) \|^2}{2\a_{\psi}} .
\end{align}
\end{definition}
We note that strongly convex functions satisfy a minor variation of the PL condition in \eqref{eq:pl}.

In our analysis, we will assume that $\Phi$ and $f$ satisfy the following properties:

\begin{assumption}[Basic assumptions for $\Phi,f$] \label{assumption0} 
	\hspace{0pt}
	\begin{itemize} [leftmargin=*]
		\setlength\itemsep{-0.0em} 
		\item $\Phi$ is twice-differentiable and $\beta_{\Phi}$-smooth.
		\item $f$ is twice-differentiable, satisfies the PL condition with $\a_f$, and $\min f(\zbf)=0$.
	\end{itemize}
\end{assumption}

Despite $f$ satisfies the PL condition, the nonconvex $\Phi$  can render $h$ nonconvex, and hence difficult to minimize in theory.
However,  we show that fast convergence of gradient descent to a global minimum can be established with appropriate initialization.

The intuition behind these assumptions is that to achieve  nonsingularity of $\Der\Phi^{*}$, we approximate $\Der\Phi^*(\wbf_0)$ at initialization and bound $\Der\Phi^*(\wbf_0) - \Der\Phi^*(\wbf_i)$ 	at iteration $i$ using the fact that $\|\wbf_0 - \wbf_i\|$ is sufficiently small by the \op. In the special case of shallow neural networks, we expect a similar argument applies even when the activation function is ReLU.  Adapting our analysis for such extensions is an interesting area of future work.

\section{Gradient flow}\label{sec:flow}
In this section, we consider gradient flow, which can be viewed as the limit of gradient descent for infinitesimally small learning rates. Inspired by the analysis of gradient flow, we provide an upper bound on the length of the trajectory traversed by gradient descent iterates and then find a sufficient condition in terms of initialization to establish its convergence to a global minimum. We focus on gradient descent in Section~\ref{sec:gd}.

Let $t\geq 0$ and consider the gradient flow, which is initialized at $\wbf_0 \in\R^{d}$ and traverses the curve
$\g:\R_+ \rightarrow \R^{d}$, given by
\begin{align}\label{eq:grad-flow}
 \dot{\g}(t)  = \frac{\der \g(t)}{\der t}= - \Der h(\g(t))
\end{align}
where $\g(0) = \wbf_0$.

We now calculate the length of the curve $\g$. 
Suppose that $\Der\Phi^*(\wbf_0)$ is $(\mu_\Phi,\nu_\Phi)$-near-isometry. Using Lemma~\ref{lem:Phi-singular-bound}, in the following lemma, we control the length inside of $\ball{(\wbf_0, \rho_{\Phi})}$. See Appendix~\ref{app:lem:length-gf} for the proof.

\blm\label{lem:length-gf} Let  $t\geq 0$ and let  $\ell(t)$ denote the length of the curve $\g$ in (\ref{eq:grad-flow}), restricted to the interval $[0,t]$. Let $t_\Phi \in (0, \infty]$ be the smallest value such that $\g(t_\Phi) \notin \ball(\wbf_0, \rho_{\Phi})$. Suppose $\Der\Phi^*(\wbf_0)$ is $(\mu_\Phi,\nu_\Phi)$-near-isometry.  Then, for all $t\le  t_\Phi$, we have
	\begin{align}\nn
	\ell(t) = O \left(\frac{\nu_\Phi \sqrt{h(\wbf_0)}}{\mu_\Phi^2 \sqrt{\a_f}}\right).
	\end{align}
\elm

Lemma~\ref{lem:length-gf} implies that if the objective value at initialization,  $h(\wbf_0)$, is sufficiently small, then we can localize gradient flows to a region around $\wbf_0$. Combining with Lemma~\ref{lem:Phi-singular-bound}, we show that the limit point of gradient flow is a global minimum. This theorem is formally stated below.

\bth[Gradient flow]\label{thm:gf-final-result}
	Let $\wbf_0\in\R^d$. Suppose that $\Phi$ and $f$ satisfy Assumption~\ref{assumption0} and $\Der\Phi^*(\wbf_0)$ is $(\mu_\Phi,\nu_\Phi)$-near-isometry. If  $\wbf_0$ satisfies
	\begin{align}\label{eq:key-assump-equiv}
	h(\wbf_0) = O\left( \frac{\a_f \mu_\Phi^6 }{\beta_\Phi^2 \nu_\Phi^2 }\right), 
	\end{align}
	then the gradient flow $\g$  in (\ref{eq:grad-flow}) converges to a global minimum.
\eth
\bpr[Proof of Theorem \ref{thm:gf-final-result}]
Proper initialization in \eqref{eq:key-assump-equiv} ensures $\ell(t_\Phi) < {\rho_\Phi}$, which implies that 
\begin{align}\label{eq:length-bnd-gk}
\| \g(t_\Phi) - \wbf_0 \| & = \| \g(t_\Phi) - \g(0) \| < \rho_\Phi. 
\end{align}
Therefore, $\g(t) \in \ball(\wbf_0,\rho_\Phi)$ for all $t\ge 0$,  and the length of $\g$ is upper bounded by $\rho_\Phi$ using Lemma~\ref{lem:length-gf}.  Hence, the gradient flow $\g$ converges, \ie the limit point $\ol{\wbf}\in \R^{d}$ exists and satisfies 
\begin{align}\label{eq:limit-is-close}
\| \ol{\wbf} - \wbf_0 \| \le \rho_\Phi. 
\end{align}
Combining \eqref{eq:iso} and \eqref{eq:limit-is-close}, we have 
\begin{align}\nn
\frac{\mu_\Phi}{2} \leq \s_{\min}(\Der \Phi^*(\ol{\wbf})) \le \s_{\max}(\Der \Phi^*(\ol{\wbf})) \leq \frac{3\nu_\Phi}{2}.
\end{align}

In particular, we note that $\Der \Phi^*(\ol{\wbf})$ is  nonsingular. So we have $\Der f(\ol{\zbf}) = 0$. Since $f$ satisfies the PL condition in~\eqref{eq:pl},  $\ol{\zbf}$ is a global minimizer of $f$, and $\ol{\wbf}$ is a global minimizer of $h$ in~\eqref{eq:main}. 
\epr

\section{Gradient descent}\label{sec:gd}
We now view gradient descent as the discretization of gradient flow, and show that a similar argument as in Section~\ref{sec:flow} holds for gradient descent.

Let $\eta>0$ denote the learning rate and let $i\geq 0$. The gradient descent update rule is given by  
\begin{align}\label{eq:GD-iterates}
 \wbf_{i+1} = \wbf_i - \eta \Der h(\wbf_i).
\end{align}
To study gradient descent, in addition to the previous assumptions on $\Phi$ and $f$ for the case of gradient flow described in Theorem \ref{thm:gf-final-result}, we also assume that $f$ is smooth, \ie there exists $\beta_f\geq 0$ such that, for all $\zbf,\zbf'\in \R^{\tilde d}$, we have  

\begin{align}
f(\zbf) - f(\zbf')  \le \langle \zbf-\zbf', \nabla f(\zbf') \rangle + \frac{\beta_f}{2} \| \zbf-\zbf'\|^2.\nn
\end{align}

Smoothness of $f$ allows safe discretization of gradient flow without deviating too much from its trajectory. The following result is the analogue of Theorem~\ref{thm:gf-final-result} for gradient descent; see Appendix~\ref{app:thm:gd-final-result} for the proof. 

\bth[Gradient descent]\label{thm:gd-final-result}
Let $\wbf_0\in \R^{d}$. Suppose that $\Phi$ and $f$ satisfy Assumption~\ref{assumption0},  $f$ is $\beta_f$- smooth, and $\Der \Phi^*(\wbf_0)$ is $(\mu_\Phi,\nu_\Phi)$-near-isometry.  Suppose that gradient descent is executed with sufficiently small learning rate 
\begin{align}\label{eq:step-size-prop-gd}
	\eta  = O\left(\frac{1}{\b_\Phi \|\Der f(\Phi(\wbf_0)\| + \b_f\mu_\Phi^2+\b_f \nu_\Phi^2}\right), 
\end{align} and $\wbf_0$ satisfies \eqref{eq:key-assump-equiv}.


Then the sequence of iterates $\{\wbf_i\}_{i\ge 0}$ converges to a global minimum of $h$  exponentially fast.

In addition, the rate of convergence is given by
\begin{align}\label{eq:conv_rate}
	h(\wbf_i)  \le (1-C \eta \alpha_f \mu_\Phi^2 )^i   h(\wbf_0)
\end{align} where $C$ is  a universal constant. 
\eth

To prove Theorem~\ref{thm:gd-final-result}, we first compute the length of the trajectory traversed by gradient descent iterates.  We then use the smoothness of $f$ and follow the descent inequality to lower bound $f(\zbf_i)  - f(\zbf_{i+1})$. Finally, we  compute the local Lipschitz constant of $f$. 

\begin{remark} The idea of initializing a nonconvex problem close to a global minimum has a long history in nonconvex optimization, particularly in matrix factorization; see ~\citep{chi2018nonconvex} and references therein. The observation that the length of the learning trajectory is short in the \op regime has a precedent in~\citep{du2018gradient,  oymak2018overparameterized}. From an algorithmic perspective, the idea of linearizing $\Phi$ when minimizing $h=f\circ \Phi$ is studied in nonlinear regression and  the Gauss-Newton method~\citep{nocedal2006numerical}. 
\end{remark}
In order to apply Theorem~\ref{thm:gd-final-result}, the key step is to verify that $h(\wbf_0)$ satisfies~\eqref{eq:key-assump-equiv}. In Section~\ref{sec:nn}, we focus on the special case of shallow neural networks and improve the state of the art.

\section{Shallow neural networks}\label{sec:nn}
In this section, we consider the problem of training shallow neural networks with gradient descent. Our strategy is to cast this problem as a special case of problem~\eqref{eq:main} and then apply Theorem~\ref{thm:gd-final-result} to establish global convergence. We start with the formal problem statement.

\subsection{Setup, assumptions, and initialization}\label{sec:nnsetu} Consider a shallow neural network with $d_0$ inputs, one hidden layer that consists of $d_1$ hidden nodes, and $d_2$ outputs. This shallow network is specified by the map 
\begin{equation}\label{eq:shallNetMap}
\begin{split}
\R^{d_0} &\mapsto \R^{d_2} \\
\xbf \ &\mapsto V \cdot \phi(W\xbf),
\end{split}
\end{equation}
where $W \in \R^{d_1\times d_0}$, $V\in \R^{d_2\times d_1}$,  and  $\phi: \R\rightarrow \R$ is an {activation function}, which is applied entry-wise.  Let  $\xbf_i\in \R^{d_0}$ and  $y_i\in \R^{d_2}$ denote the $i$-th training data and label, respectively,  for $i\in [n]$. By concatenating the training data and their labels, we form the matrices $X\in \R^{d_0\times n}$ and $Y\in \R^{d_2\times n}$.  Let denote $\Theta  = (W,V) \in \R^{d_1\times d_0} \times \R^{d_2\times d_1 }$ and $Z = \Phi (\Theta) = V \cdot \phi(W X) \in \R^{d_2 \times n}$. The fitting problem can be cast as~\eqref{eq:main} where 
\begin{align}\label{eq:f-ex-210}
h(\Theta) = f(\Phi(\Theta)) = \| V \phi(W X) - Y\|^2.
\end{align}

\begin{remark}
We assume that the activation function $\phi:\R\rightarrow\R$ is twice-differentiable. Despite this assumption excludes the popular ReLU, it is still possible to apply our results to smooth approximations of ReLU such as the softplus or Gaussian error Linear Units (GeLU) \citep{hendrycks2016gaussian,Pyr2020}. We note that softplus~\citep{Dugas2000} or GeLU~\citep{devlin2018bert}  often achieve similar or superior performance compared to the ReLU~\citep{clevert2015fast,gulrajani2017improved,Kumar2017,kim2018memorization,Xu2015}.  
\end{remark}

\begin{definition}[Hermite norm~\citep{olver2010nist}]\label{HermitNorm}
Let $\phi: \R\ra \R$. The Hermite norm of $\phi$ is given by $ \| \phi\|_{\Hc} = \sqrt{\sum_{i=0}^\infty c_i^2}$ where $c_i$ denotes the $i$-th Hermite coefficients of $\phi$ given by: 
\begin{align}\nn
c_i= \langle \phi , q_i \rangle_{\Hc} =\frac{1}{\sqrt{2\pi}} \int \phi(x) q_i(x) \exp\l(-\frac{x^2}{2}\r)\der x
\end{align} and $q_i:\R\rightarrow\R$ is the $i$-th Hermite polynomial for $i\geq 0$.
\end{definition}


In this section,  we assume that $\phi$, $f$, and data satisfy the following properties: 
\begin{assumption}[Assumptions for shallow neural networks]\label{assumption2}
	\hspace{0pt}\\
	\vspace{-5mm} 
	\begin{itemize}[leftmargin = *]
		\setlength\itemsep{-0.0em}
		\item $\phi$ is twice-differentiable, $\phi(0) = 0$, $\sup_a |\dot{\phi}(a)| = \dot{\phi}_{\max}< \infty$,  $\sup_a |\ddot{\phi}(a)|=\ddot{\phi}_{\max} < \infty$, and $ \| \phi\|_{\Hc} <\infty$. The loss function $f$ is quadratic~\eqref{eq:f-ex-210}.
		\item $\|\xbf_i\|=1$, $\|Y\|\leq 1$,  and $\s_{\max}( V_k)  = O\l(\frac{\dot{\phi}_{\max}}{\ddot{\phi}_{\max}}\r)$ for $i\in[n]$ and $k\geq 0$. 		
			\end{itemize}
\vspace{-3mm} 
\end{assumption}


%

The assumption on $\phi$ hold for GeLU, sigmoid, and tanh.  The assumption $\phi(0)=0$ is to simplify the derivations and we suspect that it can be removed at the expense of more complicated expressions. The bounded Hermite norm is a mild assumption, which is used to obtain {an upper bound on $\s_{\m}(\phi(W^0X))$} in terms of the Hermite coefficients of $\phi$. See Appendix~\ref{sec:whereHermiteAppears0} for details. The assumption on the data is fairly mild and standard in the \op literature as we can always normalize the data \citep{li2018learning, ji2019polylogarithmic}.   Similar boundedness assumptions to the last  assumption are commonly used in nonconvex optimization to guarantee convergence~\citep{FabianADMM}.  Moreover, such a bound naturally holds by applying a projection step to the gradient descent update rule, which we plan to adopt as a future work. 

\paragraph{Initialization.} We first consider the initialization scheme: 
\begin{align}
W_0 \ {\sim} \Nc(0,\o_1^2),
\quad
V_0 \ {\sim} \Nc\left(0,\o_2^2\right).
\label{eq:init-ex2-1}
\end{align}
In Section~\ref{sec:experiment}, we study the implications of our initialization and show how to possibly avoid lazy training by varying $(\omega_1, \omega_2)$. 


\subsection{Main results for shallow neural networks}
For shallow networks as described above, we verify in Appendix~\ref{app:lem:estimate_variable} that the key conditions in Lemma~\ref{lem:Phi-singular-bound} hold with high probability. Combining with Theorem~\ref{thm:gd-final-result}, we establish the global convergence guarantees. The proof in Appendix~\ref{sec:ex-2-full} uses standard tools from random matrix theory to control the random variables involved with initialization. We first estimate variables $\mu_{\Phi}, \nu_{\Phi}$ defined in Definition~\ref{cnd:iso-jac} and $\b_{\Phi}$ in  \eqref{eq:lips-cnt-DPhi} for the neural network described in Section~\ref{sec:nnsetu}.

\blm[Estimation of $\mu_{\Phi}, \nu_{\Phi},\b_\Phi$]\label{lem:estimate_variable}
	Suppose that a shallow neural network, which is constructed in Section~\ref{sec:nnsetu},  satisfies Assumption~\ref{assumption2}. Then we have
	\begin{align}\label{eq:lem8}
	\begin{split}
	\mu_\Phi &= \s_{\min}(\phi( W_0 X)), \\
	\nu_\Phi & = \dot{\phi}_{\max} \s_{\max}(X) \s_{\max}(V_0) + \s_{\max}(\phi(W_0 X)), \\
	\b_\Phi &=  \sqrt{2}\s_{\m}(X) \l( \dot{\phi}_{\m} + \ddot{\phi}_{\m} \chi_{\m}\r)
	\end{split}
	\end{align} where $\chi_{\m}=\sup_{V}\s_{\m}(V)$. 
\elm

\begin{remark} The terms $\s_{\min}(\phi( W_0 X))$ and $\s_{\m}(\phi( W_0 X))$ in \eqref{eq:lem8} play a critical role in our analysis. In \citep{du2018gradient, song2019quadratic}, strictly positiveness of the eigenvalues of \textit{Gram matrix} is the primary tool to show the convergence. \citet{oymak2019towards} also followed a similar argument using the \textit{neural network covariance matrix}. The underlying intuition seems similar to Lemma~\ref{lem:estimate_variable}. However, the resulting bounds are different since gradient descent updates $(W, V)$ simultaneously in our problem setup, which is more realistic.
\end{remark}


By combining Lemma~\ref{lem:estimate_variable} and the results on  global  convergence of gradient descent in Section \ref{sec:gd}, we establish global convergence for shallow neural network.

\bth[Shallow network with gradient descent]\label{thm:shallow-net-ex2} Consider the shallow network described in Section~\ref{sec:nnsetu} that satisfies Assumption \ref{assumption2} and  $\tau^{r_1}|\phi(a)| \leq |\phi(\tau a)| \leq\tau^{r_2}|\phi(a)|$ for  all $a$, $0<\tau<1$, and some constants $r_1,r_2$.\footnote{The last assumption holds for popular activation functions such as sigmoid, tanh, and ELU, and can be relaxed if $\o_1=1$ in \eqref{eq:init-ex2-1}.} Suppose that $\Theta_0$ is randomly initialized as in \eqref{eq:init-ex2-1} with $\o_1$ and $\o_2$, which satisfy
\begin{align}\label{eq:init-ex2-2}
\o_1\o_2 \lesssim  
	\frac{1}{\sqrt{d_0 d_1}}, 
\end{align}
	and suppose that the hidden layer width $d_1$ satisfies
	\begin{align}
	d_1 = \tilde{\Omega}\l( \xi(\Cc_\delta, t, \phi, \{c_i\}_{i \geq 0}) \frac{\s_{\m}(X)^2 \sqrt{n}}{\s_{\min}(X^{*t})^3}\r)
	\label{eq:gd-network-size}
	\end{align}
	where $\Cc_\delta$ is a set of constants, $\xi$ is a term independent to $d_0, n$, $t$ is a constant such that $n \simeq d_0^t$, and $X^{*t} \in \R^{d_0^t \times n}$ is derived from Khatri-Rao product with its $a$-th column defined as $\vector(x_a \otimes \cdots \otimes x_a) \in \R^{ d_0^t}$. Then gradient descent converges to a global minimum exponentially fast with probability at least $1- \psi(\phi, \xi,d_0,d_1,d_2,X)$.\footnote{$\psi$ can be arbitrary small.}See Appendix \ref{sec:denoument} for the exact expressions of $\xi$ and $\psi$.
\eth 

\begin{remark} Theorem~\ref{thm:shallow-net-ex2} shows that, with sufficient degree of \op, gradient descent finds a global minimum, except with an arbitrary small probability. Note that we need two conditions for Theorem \ref{thm:shallow-net-ex2} to hold, both of which are related to the \op of the network. The condition \eqref{eq:init-ex2-2} is for the concentration of random matrices, to make $\psi$ arbitrary small, and \eqref{eq:gd-network-size} is for the locality of gradient descent.
\end{remark}

\subsection{Order analysis}\label{sec:order}
We first decompose the random matrix $\phi(X^\top W_0^\top )\phi(W_0 X)$ 
 into independent random matrices. We then apply concentration inequalities to establish an upper bound on $\s_{\m}(\phi(W_0 X))$ and a lower bound on $\s_{\min}(\phi(W_0 X))$ through the Hermite decomposition of $\phi(W_0 X)$ and note that with high probability, 
\begin{align}\nn
\sqrt{\frac{c_t^2}{t!} d_1 } \s_{\min}(X^{*t})\lesssim \s_{\min}(\phi(W_0 X))\lesssim\s_{\m}(\phi(W_0 X))\lesssim \sqrt{c_0^2dn}.
\end{align}  We also find an upper bound on $h(\Theta_0)$ at initialization. Substituting $\nu_\Phi$, $\mu_\Phi$, $\b_\Phi$ into~\eqref{eq:key-assump-equiv}, we obtain the sufficient condition in \eqref{eq:gd-network-size}.  We note that $\xi(\Cc_\delta, t, \phi, \{c_i\}_{i\geq 0})$ can be viewed as a constant w.r.t. $d_0, ~d_1$, and $n$. 
For $t=1$, it requires $n \simeq d_0$, which is not a common setting in practice.  For $t \ge 2$, we suppose that $n \simeq d_0^t$, which is the case in practice and  estimate $\s_{\max}(X) \simeq \sqrt{\frac{n}{d_0}}$ and $\s_{\min}(X^{*t}) \simeq \sqrt{\frac{n}{d_0^t}} \simeq 1$ along the lines of \citep[Section 2.1]{oymak2019towards}. Substituting $\s_{\max}(X)$ and $\s_{\min}(X^{*t})$ into  $\eqref{eq:gd-network-size}$, we have
\begin{align}
d_1 & \gtrsim  \frac{n^{\frac{3}{2}}}{d_0}.
\end{align}
Therefore, the overall \op degree becomes $d_0d_1 \simeq \tilde{\Omega}(n^{\frac{3}{2}})$, which is sufficient for gradient descent to find a global minimum at a linear rate except with an arbitrary small probability.  We note that an optimal linear scaling for the width $d_1 \simeq \tilde{O}(n)$ is sufficient when the number of input features is sufficiently large $d_0\simeq\tilde\Omega(\sqrt{n}) $,  which improves upon the results of \citep{{oymak2019towards}} by a factor of $\tilde{O}(n^\frac{1}{2})$. Furthermore, unlike \citep{oymak2019towards}, we adopt standard initialization strategies in Theorem~\ref{thm:shallow-net-ex2}.

%

\section{Lazy training and experimental evaluation}\label{sec:experiment}
    \begin{wrapfigure}{R}{0.5\textwidth}
        \vspace*{-0.7cm}
            \includegraphics[width=0.5\textwidth]{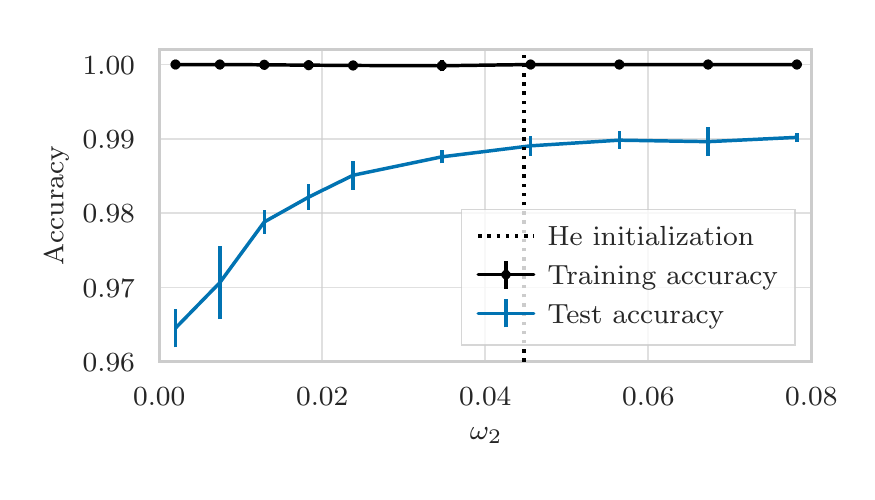}
            \caption{Training and test error on MNIST for different $\omega_2$.
		Error bars indicates the 95\% confidence interval computed over 5 independent runs. The setup details are provided in Appendix~\ref{app:exp}.}
            \label{fig:experiment}
        \vspace*{-0.4cm}
    \end{wrapfigure}
		

Following the theoretically motivated initialization in Theorem~\ref{thm:shallow-net-ex2},  we set $\o_1 \o_2 \simeq \frac{1}{\sqrt{d_0d_1}}$. This gives rise to a broad family of initialization schemes as one varies the ratio $\o_2/\o_1$.
Interestingly,  we note that popular initialization schemes such as LeCun \citep{LeCuninit} and He initialization \citep{Heinit} belong to this family. The purpose of this section is to empirically investigate the impact the choice of this ratio has on generalization of shallow networks.

To this end, we will look at the generalization error of varying initializations in the more practical setting of \text{stochastic} gradient descent (SGD). 
Specifically, we fix the product of the weight initialization $\o_1\o_2$ and then proceed by varying $\o_2$.
To ensure that perfect generalization is possible, we adopt the teacher-student setup, where, for the teacher network, we train a two-layer fully connected neural network, on MNIST~\citep{MNIST} until SGD reaches zero training error. The student networks are trained for 300 epochs to ensure convergence. The results are shown in Figure~\ref{fig:experiment}. We use mean-square loss and a smooth activation function (GeLU \citep{hendrycks2016gaussian}) for the student network to match the problem setup as closely as possible.

In Figure~\ref{fig:experiment}, we observe that while SGD achieves zero training error for every $\o_2$, as suggested by Theorem \ref{thm:shallow-net-ex2} applicable in the full batch setting, the generalization ability increases as the ratio $\omega_2/\omega_1$ grows.  It is also interesting to observe that the popular He initialization scheme corresponds to a rather balanced ratio that lies at the boundary of the well-performing region. 
In our experiments, we used He initialization to fix the value $\omega_1\omega_2$. This tendency suggests that a wide family of initialization schemes could generalize well as long as the ratio $\omega_2/\omega_1$ is not too small.

\paragraph{Comment on lazy training.}
It is important to address the so called lazy regime when generalization is of concern. Let 
\begin{align}\nn
\tilde{h}(\Theta):=h(\Theta_{0})+\langle\nabla h(\Theta_{0}), \Theta-\Theta_{0}\rangle
\end{align}
 be the linearized function of $h$ around $\Theta_{0}$ and let $\Theta_{i}$ and $\tilde{\Theta}_{i}$ denote the iterates of gradient descent at time $i$. The lazy training regime refers to the case where the training trajectory stays close to this linearization, i.e. $\|h(\Theta_{i})-\tilde{h}(\tilde{\Theta}_{i})\| \simeq 0$ for all $i$ \citep{chizat2019lazy}. 
Such a linearization occurs in infinitely wide neural networks  \citep{chizat2019lazy}, which have been shown to generalize well in some settings \citep{Arora2020Harnessing,lee2020finite}. However, in our case of subquadratic (finite) width, the lazy regime might lead to poor generalization. To gain insight on when we cannot avoid it with certainty, let us make a simple rewriting of our network assuming  $\phi$ is homogeneous:
\begin{align}\nn
\Phi(\Theta) = \alpha V \phi(WX)
\end{align}
with $V_0 \sim \Nc(0, 1)$ and $ W_0 \sim \Nc(0, 1)$ where the standard deviations are pulled out as a scaling factor $\alpha=\o_1 \o_2 \simeq 1/\sqrt{d_0d_1}$. This seems to fit into an example in \citep[Appendix A.2]{chizat2019lazy} suggesting lazy training as $d_1 \ra \infty$. However, their results require an odd activation function and infinite width, while our activation function is required not to be odd (see the proof in Appendix~\ref{sec:ex-2-full}) and our results are under subquadratic (finite) width. Instead, to study lazy training, we explicitly compute an upper bound on $\|h(\Theta_i)-\tilde{h}(\tilde{\Theta}_{i})\|$ in Appendix~\ref{app:lazy} following \citep[Theorem 2.3]{chizat2019lazy}.

It turns out that the upper bound becomes $\infty$ when $\o_1 \ll \o_2$, and it becomes zero when $\o_1 \gg \o_2$.  Our analysis suggests that shallow neural networks can avoid lazy training provided that $\o_2/\o_1 \ra \infty$. 
This analysis is corroborated by the empirical results showing that the generalization capability improves as $\o_2$ grows in Figure~\ref{fig:experiment}. On the other hand, if $\o_2/\o_1 \ra 0$, then lazy training is bound to happen asymptotically.  For details, see Appendix~\ref{app:lazy}. Finally, we note that we have not theoretically claimed
that our initialization is guaranteed to be non-lazy, since doing so would require establishing a lower bound on $\|h(\Theta_{i})-\tilde{h}(\tilde{\Theta}_{i})\|$, which is an interesting problem for future work.  Instead, our discussion above only provides a necessary condition for non-lazy training, and a sufficient condition for lazy training.

\section{Conclusions and future work}\label{sec:conc}
In this paper, we prove the linear convergence of first-order methods on subquadratically \oped two-layer neural networks with smooth activation functions. Our theoretical analysis is compatible with standard initialization strategies, which can potentially avoid lazy training. We train both layers simultaneously and achieve  a desirable subquadratic scaling on the width of the network. In particular, we note that a linear scaling for the width $d_1 \simeq \tilde{O}(n)$ is sufficient when the number of input features is sufficiently large $d_0\simeq\tilde\Omega(\sqrt{n}) $. We use tools from random matrix theory under standard assumptions on data and leverage on the assumption that the loss satisfies Polyak-{\L}ojasiewicz condition.  We carefully find an explicit upper bound and lower bound on singular values of the outputs of the first layer at initialization with high probability under general initialization. 

It is natural to ask whether we can attain similar degree of \op with nonsmooth activation functions such as ReLU.  We plan to adapt our analysis for such extensions as a future work.  While our analysis provides a necessary condition for avoiding lazy training, it is interesting to develop sufficient conditions in the future. In particular,  developing lower bounds on $\|h(\Theta_{i})-\tilde{h}(\tilde{\Theta}_{i})\|$ will be a key to fully characterize lazy training.

Finally, as a theoretical work, we do not anticipate any potential negative societal impacts of our paper. However,  the long-term impacts of our work may depend on how machine learning algorithms are used in society.

\begin{ack}
The authors would like to thank Fabian Latorre, Fanghui Liu,  and Paul Rolland for helpful discussions.

This project has received funding from the European Research Council (ERC) under the European Union's Horizon 2020 research and innovation programme (grant agreement n° 725594 - time-data). This project  was sponsored by the Department of the Navy, Office of Naval Research (ONR)  under a grant number N62909-17-1-2111. This work was supported by Hasler Foundation Program: Cyber Human Systems (project number 16066). Research was sponsored by the Army Research Office and was accomplished under Grant Number W911NF-19-1-0404.

\end{ack}

\bibliographystyle{plainnat}
\bibliography{Ref}

\appendix

\section{Proof of Lemma~\ref{lem:Phi-singular-bound}}\label{app:lem:Phi-singular-bound}
Intuitively, if $\Der \Phi^*(\wbf_0)$ is a $(\mu_\Phi,\nu_\Phi)$-near-isometry, then one would  expect $\Der \Phi^*$ to remain near-isometry for all nearby points.  Formally, let $A,~B\in R^{m\times n}$ and let singular values of a matrix are ordered such that $\s_i(A)\geq \s_j(A)$ and $\s_i(B)\geq \s_j(B)$ for $1\leq i\leq j\leq \min\{m,n\}$.  Using Weyl's inequality and for $i+j-1\leq \min\{m,n\}$, we have: 

\begin{align}\label{Weyl}
\s_{i+j-1}(A+B) \leq \s_i(A)+ \s_j(B).
\end{align}

More formally, suppose that $\wbf\in \R^{d}$ satisfies 
\begin{align}\label{eq:perturb-assump}
\|\wbf -\wbf_0 \| \leq \frac{\mu_\Phi}{2\b_\Phi } =\rho_\Phi.
\end{align}
If $\Der \Phi^*(\wbf_0)$ is $(\mu_\Phi,\nu_\Phi)$-isometry in the sense of Definition~\ref{cnd:iso-jac}, then applying Weyl's inequality~\eqref{Weyl} along with using smoothness and \eqref{eq:perturb-assump},   we have  
\begin{align}
\s_{\min}( \Der \Phi^*(\wbf)) & \ge \s_{\min}( \Der \Phi^* (\wbf_0) ) - \s_{\max}( \Der \Phi^* (\wbf) - \Der \Phi^* (\wbf_0) ) \nn\\
& \ge \mu_\Phi  - \b_\Phi \| \wbf - \wbf_0\| \nn\\
& \ge \frac{\mu_\Phi}{2}.\nn
\end{align}
Using a similar argument, we establish an upper bound $\s_{\max}(\Der \Phi^*(\wbf))$:
\begin{align}\nn
\s_{\max}( \Der \Phi^*(\wbf))\leq  \s_{\max}( \Der \Phi^*(\wbf_0))+\s_{\max}( \Der \Phi^* (\wbf) - \Der \Phi^* (\wbf_0))\leq \nu_\Phi+\frac{\mu_\Phi}{2}\leq \frac{3\nu_\Phi}{2}. 
 \end{align}

\section{Proof of Lemma~\ref{lem:length-gf}}\label{app:lem:length-gf}
Let $t\geq 0$ and denote 
\begin{align}\label{eq:shorthand-gf}
\z(t) = \Phi(\g(t))
\end{align}
so we have  
\begin{align}\label{eq:recall-h-f}
h(\g(t)) = f(\Phi(\g(t)) = f(\z(t)).
\end{align}

Taking the first-order derivative w.r.t. $t$, we have  
\begin{align}
\begin{split}
\dot{ \z}(t) & = \Der \Phi( \g(t)) \bl \dot{\g}(t) \br
\\
& = -  \Der \Phi (\g(t)) \bl \Der h( \g(t)) \br.
\label{eq:g-to-gK}
\end{split}
\end{align}
Note that we have 
\begin{align}\label{eq:derone-exp}
\begin{split}
\frac{\der h(\g(t))}{\der t} & = 
\Der h(  \g(t)) \bl \dot{\g}(t) \br\\
& = - \Der h(\g(t)) \bl \Der h(\g(t)) \br\\
& = - \| \Der h(\g(t)) \|^2.
\end{split}
\end{align}

Length of the segment of the curve $\g_K$ restricted to the interval $[0,t]$ is given by 
\begin{align}\label{eq:length-lK}
\begin{split}
\ell(t) & = \int_0^{t} \| \dot{\g}(\t) \| \der \t\\
& = \int_0^{t} \| \Der h(\g(\t) )\| \der \t \\
& \le \int_0^{t} \s_{\max}(\Der \Phi^*(\g(\t)) \cdot \| \Der f(\z(\t)) \| \der \t\\
& \lesssim \nu_\Phi \int_0^{t} \| \Der f(\z(\t))\| \der \t. 
\end{split}
\end{align}

To control the norm in the last line of \eqref{eq:length-lK}, we note that 
\begin{align}
\begin{split}
- \frac{\der \sqrt{f(\z(\t)) - f(\z(t))}}{\der \t} 
& = - \frac{ \frac{\der f(\z(\t))}{\der \t } }{2\sqrt{f(\z(\t))- f(\z(t))}}\\ 
& = -\frac{\langle \Der f(\z(\t)), \dot{\z}(\t)\rangle }{2 \sqrt{f(\z(\t))-f(\z(t))}}\\
& = \frac{\l\langle \Der f(\z(\t)), \Der \Phi(\g(\t)) \bl \Der h(\g(\t)) \br\r \rangle }{2 \sqrt{f(\z(\tau))-f(\z(t))}} \\
&=  \frac{\| \Der h(\g(\t)) \|^2}{2 \sqrt{f(\z(\t))- f(\z(t))}}\\
& \ge \frac{\s_{\min}^2( \Der \Phi^*(\g(\t)))  \cdot \| \Der f(\z(\tau))\|^2 }{2 \sqrt{f(\z(\t))- f(\z(t))}}\\
& \gtrsim \frac{\mu_\Phi^2 \cdot \| \Der f(\z(\t))\|^2 }{ \sqrt{f(\z(\t))- f(\z(t))}}\\
& \gtrsim \frac{\sqrt{ \a_f}  \mu_\Phi^2 \cdot \| \Der f(\z(\t))\|^2 }{\| \Der f(\z(\t)) \|}\\
& = \sqrt{ \a_f}  \mu_\Phi^2 \cdot \| \Der f(\z(\t))\|,
\label{eq:grad-to-val}
\end{split}
\end{align}
provided that the denominators are nonzero. 
Substituting \eqref{eq:grad-to-val} into \eqref{eq:length-lK},  the desired length is bounded by  
\begin{align}
\ell(t) & \lesssim \nu_\Phi \int_0^{t} \| \Der f(\z(\t))\| \der \t \nn\\
& \lesssim - \frac{ \nu_\Phi}{\mu_\Phi^2 \sqrt{\a_f}} \int_0^{t} \frac{\der \sqrt{f(\z(\t))- f(\z(t))}}{\der \t} \der \t
\nn\\
& = \frac{ \nu_\Phi}{\mu_\Phi^2 \sqrt{\a_f}} \l(\sqrt{f(\z(0))} - \sqrt{f(\z(t))} \r) \nn\\
& \le 
\frac{ \nu_\Phi  \sqrt{f(\z(0))} }{\mu_\Phi^2 \sqrt{\a_f}}
\nn\\
& = \frac{ \nu_\Phi \sqrt{h(\g(0))}  }{\mu_\Phi^2 \sqrt{\a_f}} 
\nn\\
& =  \frac{ \nu_\Phi \sqrt{ h(\wbf_0)} }{\mu_\Phi^2 \sqrt{\a_f}} ,\nn
\end{align}
which completes the proof of Lemma \ref{lem:length-gf}.

\section{Proof of Theorem~\ref{thm:gd-final-result}}\label{app:thm:gd-final-result}
The proof is along the lines of Theorem~\ref{thm:gf-final-result}. We first compute the length of the trajectory traversed by gradient descent iterates. Formally, let $I$ denote the first iteration such that $\wbf_I\notin \ball(\wbf_0,\rho_\Phi)$. The length of the trajectory traced by $\{\wbf_i\}_{i=0}^I$ is upper bounded by  
\begin{align}\label{eq:length-step1-gd}
\begin{split}
\ell(I) & := \sum_{i=0}^{I-1} \| \wbf_{i+1}- \wbf_i\| \\
& = \eta \sum_{i=0}^{I-1} \| \Der h(\wbf_i) \| 
\\
& \lesssim \eta \nu_\Phi \sum_{i=0}^{I-1} \|\Der f(\zbf_i)\|.
\end{split}
\end{align}

This following lemma is useful for our proof. 
\blm\label{lm:Phi_var} Suppose $\ubf,\vbf\in\ball(\wbf_0,\rho_\Phi)$. Then we have
$\|\Phi(\ubf)-\Phi(\vbf)\|\leq \frac{3\nu_\Phi}{2}\|\ubf-\vbf\|$.
\bpr
Using Lemma~\ref{lem:Phi-singular-bound}, we establish a bound on $\|\Phi(\ubf)-\Phi(\vbf)\|$: 
\begin{align}
\|\Phi(\ubf)-\Phi(\vbf)\|&=\Big\|\int_{0}^1\Der\Phi(\vbf+t(\ubf-\vbf))(\ubf-\vbf)\der t\Big\|\nn\\
&\leq \int_{0}^1\|\Der\Phi(\vbf+t(\ubf-\vbf))(\ubf-\vbf)\|\der t\nn\\
&\leq \frac{3\nu_\Phi}{2}\|\ubf-\vbf\|.\nn
\end{align}
\epr
\elm

Let $i\leq I-2$. To control the upper bound in~\eqref{eq:length-step1-gd}, we use the smoothness of $f$ and Lemma~\ref{lm:Phi_var} to obtain a standard ``descent inequality'' as: 
\begin{align}
f(\zbf_i)  - f(\zbf_{i+1})& \geq \langle \zbf_i - \zbf_{i+1}, \Der f(\zbf_i) \rangle - \frac{\b_f}{2}\| \zbf_{i+1}-\zbf_i \|^2\nn\\
& =\langle \Phi(\wbf_i) - \Phi(\wbf_{i+1}), \Der f(\zbf_i) \rangle - \frac{\b_f}{2}\| \Phi(\wbf_{i+1})-\Phi(\wbf_i) \|^2\nn\\ 
& = \langle \Der \Phi(\wbf_i) \bl \wbf_i - \wbf_{i+1} \br, \Der f(\zbf_i) \rangle - \frac{\b_f}{2}\| \Phi(\wbf_{i+1})-\Phi(\wbf_i) \|^2\nn\\
& \qquad -\langle \Phi(\wbf_{i+1})- \Phi(\wbf_i) - \Der \Phi(\wbf_i) \bl \wbf_{i+1}-\wbf_{i} \br   , \Der f(\zbf_i) \rangle \nn\\
& \ge \langle \Der \Phi(\wbf_i) \bl \wbf_i - \wbf_{i+1} \br, \Der f(\zbf_i) \rangle - \frac{\b_f}{2}\| \Phi(\wbf_{i+1})-\Phi(\wbf_i) \|^2\nn\\
& \qquad - \frac{\b_\Phi}{2}\|\wbf_{i+1}-\wbf_{i}\|^2 \|\Der f(\zbf_i)\| \nn\\
& \ge  \langle \Der \Phi(\wbf_i) \bl \wbf_i - \wbf_{i+1} \br, \Der f(\zbf_i) \rangle 
- \frac{1}{2} \| \wbf_{i+1} - \wbf_i \|^2 \l(  \b_\Phi \|\Der f(\zbf_i) \| + \frac{9 \b_f \nu_\Phi^2}{4} \r)
\nn\\
& =  \eta \langle \Der \Phi(\wbf_i) \bl \Der h(\wbf_i) \br, \Der f(\zbf_i) \rangle 
- \frac{\eta^2}{2} \|\Der h(\wbf_i) \|^2 \l( \b_\Phi \|\Der f(\zbf_i) \| + \frac{9\b_f  \nu_\Phi^2}{4} \r) \nn\\
& =  \eta \| \Der h(\wbf_i) \|^2 - \frac{\eta^2 }{2} \|\Der h(\wbf_i) \|^2  \l( \b_\Phi \|\Der f(\zbf_i) \| + \frac{9 \b_f \nu_\Phi^2}{4} \r) 
\nn\\
& =  \eta \| \Der h(\wbf_i) \|^2\l(1 - \frac{\eta \b_\Phi  \| \Der f(\zbf_i) \|}{2} - \frac{9\eta \b_f \nu_\Phi^2}{8} \r)
\nn\\
& \gtrsim  \eta\mu_\Phi^2 \| \Der f(\zbf_i)\|^2\quad\text{(chain rule and Lemma~\ref{lem:Phi-singular-bound})}\nn
\end{align}
where the fourth inequality holds since $\|\Phi(\abf)-\Phi(\bbf)-\Der \Phi(\bbf)(\abf-\bbf)\|\leq \frac{\b_\Phi}{2}\|\bbf-\abf\|^2$ for $\b_\Phi$-smooth $\Phi$, and the last line holds provided that  $\eta$ satisfies: 
\begin{align}\label{eq:bnd-step-size}
\eta \lesssim \frac{1}{ \b_\Phi \max_i \|\Der f(\zbf_i) \|+ \b_f \nu_\Phi^2} .
\end{align}

We now use the bound above to find an upper bound on $\sqrt{f(\zbf_i)-f(\zbf_{I-1})} - \sqrt{f(\zbf_{i+1})- f(\zbf_{I-1})}$: 
\begin{align}\label{eq:diff-in-roots-gd}
\begin{split}
\sqrt{f(\zbf_i)-f(\zbf_{I-1})} - \sqrt{f(\zbf_{i+1})- f(\zbf_{I-1})}
& = 
\frac{f(\zbf_i) - f(\zbf_{i+1})}{\sqrt{f(\zbf_{i})-f(\zbf_{I-1})} + \sqrt{f(\zbf_{i+1})-f(\zbf_{I-1})}} \\
& \gtrsim \frac{ \eta \mu_\Phi^2 \| \Der f(\zbf_i) \|^2 }{\sqrt{f(\zbf_{i})-f(\zbf_{I-1})} + \sqrt{f(\zbf_{i+1})-f(\zbf_{I-1})}} 
\\
& \ge \frac{\eta\mu_\Phi^2 \| \Der f(\zbf_i)\|^2}{2\sqrt{f(\zbf_i)-f(\zbf_{I-1})}}
\\
& \ge \frac{ \eta \sqrt{\a_f} \mu_\Phi^2 \| \Der f(\zbf_i) \|^2 }{ \sqrt{2} \| \Der f(\zbf_i) \|  } 
\\ 
& = \frac{ \eta \sqrt{\a_f} \mu_\Phi^2  }{ \sqrt{2}   } \| \Der f(\zbf_i) \| . 
\end{split}
\end{align}
{ Substituting \eqref{eq:diff-in-roots-gd} into \eqref{eq:length-step1-gd}, we have 
\begin{align}\label{eq:trabound}
\begin{split}
\ell(I) & \lesssim \eta\nu_\Phi \sum_{i=0}^{I-1} \| \Der f(\zbf_i)\| 
\\
& \lesssim \frac{\nu_\Phi}{ \sqrt{\a_f}\mu_\Phi^2} \sum_{i=0}^{I-2} \l(\sqrt{f(\zbf_i) - f(\zbf_{I-1}) } - \sqrt{f(\zbf_{i+1}) - f(\zbf_{I-1}) } \r)+\eta\nu_\Phi \| \Der f(\zbf_{I-1})\|
\\
& \lesssim \frac{\nu_\Phi}{ \sqrt{\a_f}\mu_\Phi^2} \sqrt{f(\zbf_0) - f(\zbf_{I-1}) }+\eta\nu_\Phi \| \Der f(\zbf_{I-1})\|\\
& \le \frac{\nu_\Phi \sqrt{f(\zbf_0)}  }{ \sqrt{\a_f} \mu_\Phi^2} +\eta\nu_\Phi \| \Der f(\zbf_{I-1})\|.
\end{split}
\end{align}
Note that 
\begin{align}
f(\zbf_0) = h(\wbf_0) \lesssim \frac{\a_f \mu_\Phi^6}{ \b_{\Phi}^2 \nu_\Phi^2}\nn
\end{align} {and scaling down the learning rate sufficiently to control the second term in the upper bound}
ensure that 
\begin{align}
\ell(I) \le \frac{\rho_\Phi}{2} = \frac{\mu_\Phi}{4\b_{\Phi}}.\nn
\end{align}
}
Hence, the gradient descent iterates satisfy:  
\begin{align}
\{\wbf_i\}_{i\ge 0} \in \text{ball}(\wbf_0,\rho_\Phi),\nn
\end{align}
which implies that the limit $\ol{\wbf}$ exists and is  globally optimal. In the following, we simplify the expression for $\eta$ in~\eqref{eq:bnd-step-size}. Since the iterates of gradient flow remain within a ball of radius $\rho_\Phi$, we can compute the local Lipschitz constant of $f$ as

\begin{align}\label{eq:local-lips-gd}
\begin{split}
\max_i \| \Der f(\zbf_i) \| & \le \|\Der f(\zbf_0)\|+ \max_i \| \Der f(\zbf_i)-\Der f(\zbf_0) \| \\
& \le \| \Der f(\zbf_0) \| + \b_f \max_i \|\zbf_i - \zbf_0\| 
\\
& = \| \Der f(\zbf_0) \| + \b_f \max_i \| \Phi(\wbf_i)- \Phi(\wbf_0) \| 
\\
& = \|\Der f(\zbf_0)\| + \frac{3\b_f\nu_\Phi}{2}  \max_i \|\wbf_i - \wbf_0\| 
\\
& \le  \|\Der f(\zbf_0)\| + \frac{3\b_f\nu_\Phi}{2} \cdot \rho_\Phi 
\\
& =  \|\Der f(\zbf_0)\| + \frac{3\b_f\mu_\Phi\nu_\Phi}{4\b_{\Phi}}.
\end{split}
\end{align}

Substituting~\eqref{eq:local-lips-gd} into \eqref{eq:bnd-step-size}, an upper bound on $\eta$ is given by
\begin{align}
\eta \lesssim  \frac{1}{  \b_\Phi \|\Der f(\zbf_0)\| + \b_f\mu_\Phi\nu_\Phi +\b_f \nu_\Phi^2}\leq \frac{1}{  \b_\Phi \|\Der f(\zbf_0)\| + \b_f\mu_\Phi^2+\b_f \nu_\Phi^2}
\end{align} where the last inequality holds since $\mu_\Phi\leq\nu_\Phi$.

Finally, using \eqref{eq:pl}, we prove the linear convergence to the limit point $\ol{\wbf}$: 
\begin{align}
h(\wbf_{i+1})  & = h(\wbf_{i+1}) - h(\wbf_i) + h(\wbf_i)
\nn\\
& = f(\zbf_{i+1}) - f(\zbf_i) + h(\wbf_i)
\nn\\
& \le- C  \eta\mu_\Phi^2 \| \Der f(\zbf_i) \|^2 + h(\wbf_i) 
\nn\\
& \le (1- C \eta\a_f \mu_\Phi^2) h(\wbf_i) 
\end{align}
where $C$ is a universal constant. This completes the proof of Theorem~\ref{thm:gd-final-result}. 

\section{Proof of Lemma~\ref{lem:estimate_variable}}\label{app:lem:estimate_variable}
We first obtain the expression for adjoint operator $\Der \Phi^*(\Theta):\R^{d_2\times n} \ra \R^{d_1\times d_0}\times  \R^{d_2\times d_1} $. Let $\Delta_W\in\R^{d_1\times d_0}$, $\Delta_V\in\R^{d_2\times d_1}$, and $\Delta\in \R^{d_2\times n}$. We expand $\Phi$ as follow:

\begin{align}\label{Phiexpand}
\begin{split}
\Phi(W+\Delta_W,V)&\approx \Phi(W,V)+ \Der_{W}\Phi(\Delta_W),\\
\Phi(W,V+\Delta_V)&\approx \Phi(W,V)+ \Der_{V}\Phi(\Delta_V)
\end{split}
\end{align} where 
\begin{align}\nn
\Der_{W}\Phi(\Delta_W)=V\l(\dot{\phi}(W  X)\odot \Delta_W X\r),\quad \Der_{V}\Phi(\Delta_V)= \Delta_V{\phi}(W  X),
\end{align} $\odot$ stands for the Hadamard (entry-wise) product, and $ \dot{\phi}(W X)$ is the derivative of $\phi$ calculated at each entry of the matrix $W X$.  The operator $\Der \Phi(\Theta)$ is given by $(\Delta_W,\Delta_V)\ra \Der_{W}\Phi(\Delta_W)+\Der_{V}\Phi(\Delta_V)$.

Using the cyclic property of the $\tr$ operator and $\tr\big((A\odot B)C\big)=\tr\big((A\odot C^\top)B^\top\big)$, we have 
\begin{align}\label{DirDerDot}
\begin{split}
\l\langle\Delta, \Der_{W}\Phi(\Delta_W)\r\rangle&= \l\langle \l(\dot\phi (W X)\odot {V}^\top\Delta\r)X^\top, \Delta_W\r\rangle,\\
\l\langle\Delta, \Der_{V}\Phi(\Delta_V)\r\rangle&= \l\langle \Delta_V,\Delta\phi\l(X^\top {W}^\top\r)\r\rangle.
\end{split}
\end{align}

Substituting~\eqref{DirDerDot}, the adjoint operator is given by 
\begin{align}\label{eq:derPhiStar-ex2}
\Der \Phi^*(\Theta): \Delta \ra \l( \l(\dot\phi (W X)\odot {V}^\top\Delta\r)X^\top,  \Delta\phi\l(X^\top {W}^\top\r) \r).
\end{align}

Suppose that there exist $\dot{\phi}_{\m},\ddot{\phi}_{\m}<\infty$ such that 
\begin{align}\label{eq:props-phi}
\sup_a |\dot{\phi}(a)| \le \dot{\phi}_{\m} ,
\quad 
\sup_a |\ddot{\phi}(a)| \le \ddot{\phi}_{\m}.
\end{align}

\blm\label{lm:Fnorm}
Let $A\in\R^{m\times n}$ and $B\in\R^{n\times k}$. Then, we have 
\begin{align}\nn
\s_{\min}(A)\|B\|\leq \|AB\|\leq \s_{\m}(A)\|B\|.   
\end{align}
\elm

Using Lemma~\ref{lm:Fnorm} and triangular inequality, we note that 
\begin{align}\label{eq:dphi-ubnd-delta}
\begin{split}
\| \Der \Phi^*(\Theta,\Delta)\| & \le \l\|    \l( \dot{\phi}(W  X) \odot ( {V}^\top \Delta ) \r) X^\top \r\| + \l\| \Delta   \phi(X^\top {W}^\top)  \r\|\\
& \le \dot{\phi}_{\m} \s_{\m}(X)  \s_{\m}(V)  \|\Delta\| 
+ \s_{\max}(\phi( W X))  \|\Delta\|.
\end{split}
\end{align}

Similarly, we have this lower bound:
\begin{align}\label{eq:dphi-lbnd-delta}
\| \Der \Phi^*(\Theta,\Delta)\| & \ge \s_{\min}(\phi(W X))  \|\Delta\|.
\end{align}

Substituting $\Theta_0 = (W_0,V_0)$ into \eqref{eq:dphi-ubnd-delta} and \eqref{eq:dphi-lbnd-delta},  $\mu_\Phi$ and $\nu_\Phi$ are given by:  
\begin{align}\label{eq:mu-nu-Ex2}
\begin{split}
\s_{\m}(\Der \Phi^*(\Theta_0)) &\le  \dot{\phi}_{\m} \s_{\m}(X)  \s_{\m}(V_0) 
+ \s_{\m}(\phi(W_0 X))  =: \nu_\Phi ,\\
\s_{\min}(\Der \Phi^*(\Theta_0))  & \ge  \s_{\min}(\phi(W_0 X)) =: \mu_\Phi.
\end{split}
\end{align}



In the following, we find the smoothness parameter $\b_\Phi$ in \eqref{eq:lips-cnt-DPhi}. Let $\Theta,\hat \Theta \in \R^{d_1 \times d_0} \times \R^{d_2\times d_1}$. We note that $\| \Der \Phi (\Theta,\Delta) - \Der \Phi (\hat \Theta,\Delta) \| \le U_1+U_2$ where 
\begin{align}\label{eq:lips-of-derPhi}
\begin{split}
U_1&=\| V (\dot{\phi}({W}^\top X )\odot (\Delta_W^\top X)) - \hat V(\dot{\phi}({\hat W}^{\top} X )\odot (\Delta_W^\top X))\|
\\ U_2&= \| \Delta_V \phi({W}^\top X) - \Delta_V \phi({\hat W}^\top X) \| .
\end{split} 
\end{align}

Let us denote 
\begin{align}\label{eq:Xball-limited}
\s_{\m}({\hat V}) \le \chi_{\m}.
\end{align}

An upper bound on $U_1$ in \eqref{eq:lips-of-derPhi} is given by: 
\begin{align}
U_1
& \le \| (V - {\hat V}) ( \dot{\phi}({W}^\top X) \odot (\Delta_W^\top X) ) \| + 
\| {\hat V}  ( \dot{\phi}({W}^\top X) \odot (\Delta_W^\top X) - {\hat V} \dot{\phi}(\hat W^\top X) \odot (\Delta_W^\top X) )\|\nn\\
&\le  \dot{\phi}_{\m} \s_{\m}(X) \| V - {\hat V}\|   \| \Delta_W\| 
+\s_{\m}(X) \s_{\m}({\hat V})  \|\dot{\phi}({W}^\top X) - \dot{\phi}(\hat W^{\top} X) \|_\infty \| \Delta_W\|\nn\\  
&\le   \dot{\phi}_{\m} \s_{\m}(X) \| V - {\hat V}\|   \| \Delta_W\| 
+ \ddot{\phi}_{\m} \s_{\m}(X) \| X\|_{\infty}  \s_{\m}({\hat V}) \|W-{\hat W} \| \|\Delta_W\|\nn\\
&  \le  \dot{\phi}_{\m} \s_{\m}(X) \| V - {\hat V}\|   \| \Delta_W\| + \ddot{\phi}_{\m} \chi_{\m} \s_{\m}(X)    \|W-{\hat W} \| \| \Delta_W\|\nn.
\end{align}

An upper bound on $U_2$ in \eqref{eq:lips-of-derPhi} is given by: \begin{align}\nn
U_2 \le \dot{\phi}_{\m} \s_{\m}( X)  \| W - {\hat W} \| \| \Delta_V \|. 
\end{align}
Substituting the upper bounds on $U_1$ and $U_2$, an upper bound on  $\s_{\m}( \Der \Phi (\Theta) - \Der \Phi (\hat \Theta))$ is given by
\begin{align}
\s_{\m}(\Der \Phi (\Theta) - \Der \Phi (\hat \Theta) ) &\le 
\s_{\m}(X)\l( \dot{\phi}_{\m}+ \ddot{\phi}_{\m} \chi_{\m}  \r) \| W - {\hat W} \|
+ \s_{\m}(X)\dot{\phi}_{\m} \|V - {\hat V}\|
\nn\\
& \le \sqrt{2}\s_{\m}(X) \l( \dot{\phi}_{\m} + \ddot{\phi}_{\m} \chi_{\m}   \r) \| \Theta - \hat \Theta\|\nn
\end{align} where the last inequality holds since  
\begin{align}\nn
\| W - \hat W\| + \| V - \hat V\| & \le  \sqrt{2} \sqrt{\| W - \hat W\|^2 + \| V - \hat V\|^2}.  
\end{align}

Finally, $\b_\Phi$ in \eqref{eq:lips-cnt-DPhi} is given by  
\begin{align}\label{eq:bPhi-Ex2}
\b_\Phi =  \sqrt{2}\s_{\m}(X) \l( \dot{\phi}_{\m} + \ddot{\phi}_{\m} \chi_{\m}\r). 
\end{align}

\section{Proof of Theorem \ref{thm:shallow-net-ex2}}\label{sec:ex-2-full}

This is our setup: $\min_{\Theta\in R^{d_1\times d_0}\times R^{d_2\times d_1}} h(\Theta)$ where 
\begin{align}\nn
h(\Theta) =  \|V \phi(WX ) - Y\|^2.
\end{align} Note that $\a_f=\b_f=2$.

Suppose that there exists $\chi_{\max}<\infty$ such that, for all $i\ge 0$,  we have 
\begin{align}
\s_{\m}( V_i ) \le \chi_{\m}.\nn
\end{align}
The details of $\chi_{\max}$ later will be provided in Section~\ref{sec:denoument}.

In Lemma~\ref{lem:estimate_variable}, we have shown that 
	\begin{align}\nn
	\begin{split}
	\mu_\Phi &= \s_{\min}(\phi( W_0 X)), \\
	\nu_\Phi & = \dot{\phi}_{\max} \s_{\max}(X) \s_{\max}(V_0) + \s_{\max}(\phi(W_0 X)), \\
	\b_\Phi &=  \sqrt{2}\s_{\m}(X) \l( \dot{\phi}_{\m} + \ddot{\phi}_{\m} \chi_{\m}\r).
	\end{split}
	\end{align}

In order to apply Theorem~Theorem~\ref{thm:gd-final-result}, we now establish high-probability bounds on random quantities $\mu_\Phi,\nu_\Phi$, and $h(\Theta_0)$ given the initialization in \eqref{eq:init-ex2-1}. 

\subsection{Estimating $\mu_\Phi,\nu_\Phi$} \label{sec:whereHermiteAppears0}
We now  estimate the random quantities $\mu_\Phi,\nu_\Phi$ in our neural network setting. They key quantities to estimate are $\s_{\min}(\phi( W_0 X))$ and $\s_{\m}(\phi( W_0 X))$. To that end, we consider Hermite decomposition of the activation function $\phi$.

We start with the basic definition of Hermite polynomial and its properties. Let $i\ge 0$ and let $q_i:\R\ra\R$ denote the $i$-th Hermite polynomial. Note that $q_i$'s form an orthogonal basis for the Hilbert space of functions.:
\begin{align}\nn
\Hc = \l\{ u:\R\ra\R \, |\,  \int u^2(x) \exp\l(-\frac{x^2}{2}\r) < \infty\r\}, 
\end{align} which is equipped with the inner product 
\begin{align}\nn
\langle u, v \rangle_{\Hc} =\frac{1}{\sqrt{2\pi}} \int u(x) v(x) \exp\l(-\frac{x^2}{2}\r) \der x  
\end{align} for $u,v\in \Hc$.
We consider probabilist's convention of Hermite polynomial. Specifically, for $i,j\ge 0$, we have 
\begin{align}\label{eq:orth}
\langle q_i , q_j \rangle_{\Hc} = 
\begin{cases}
i! & i=j,\\
0 & i\ne j.
\end{cases}
\end{align}

Using the above orthogonal basis to decompose  $\phi( W_0 X)$, we have
\begin{align}\label{eq:hermite-decomp}
\phi( W_0 X) & = \sum_{i=0}^\infty \frac{c_i}{i!} \cdot q_i( W_0 X )
\end{align}
where $c_i = \langle \phi, q_i \rangle_{\Hc}$ and each matrix $q_i(W_0 X)\in \R^{d_1\times n}$ is formed by applying $q_i$ entry-wise to the matrix $W_0 X$.  Let us denote
\begin{align}\nn
M_0:= \phi(X^\top W_0^\top)\phi( W_0 X).
\end{align}

Let $0<\tau<1$.  Suppose there are constants $r_1,r_2$ such that $\tau^{r_1}|\phi(a)| \leq |\phi(\tau a)| \leq\tau^{r_2}|\phi(a)|$ for all $a$.  In the following, we first obtain $\E[\tilde M_0]=\E[\phi(X^\top \tilde W_0^\top)\phi(\tilde W_0 X)]$ with $\tilde W_0\sim \Nc(0,1)$ and then obtain a lower bound on $\s_{\min}( \E[M_0] )$ and an upper bound on $\s_{\min}( \E[M_0] )$ by scaling the variance. 


Applying Hermite decomposition~\eqref{eq:hermite-decomp} and taking expectation,  we have  
\begin{align}\label{eq:prod-of-decomps-Ex2}
\begin{split}
\E[\tilde M_0] &= \E\l[ \phi(X^\top \tilde W_0^\top) \phi (\tilde W_0 X) \r] \\
& = \sum_{i,j=0}^\infty \frac{c_i c_j}{i!j!} \E[ q_i(X^\top \tilde W_0^\top) q_j( \tilde W_0 X)] 
\end{split}
\end{align}
where the expectation is w.r.t. the random matrix $\tilde W_0$. Let $\xbf_a\in \R^{d_0}$ denote the $a$-th column of the training data $X$. Each summand in~\eqref{eq:prod-of-decomps-Ex2} is an $n\times n$ matrix where 
\begin{align}\label{eq:entry}
\l[ \E[ q_i(X^\top \tilde W_0^\top) q_j(\tilde W_0 X)] \r]_{a,b} = & \sum_{c=1}^{d_1} \E \l[ q_i(\xbf_a^\top \tilde W_{0,c,\ra} )q_j(\tilde W_{0,c,\ra}^{\top} \xbf_b ) \r], 
\end{align}
where  $\tilde W_{0,c,\ra}$ is the $c$-th row of $\tilde W_0$ for $a,b\in[n]$.

In summand on the RHS of \eqref{eq:entry}, we note that there is a linear combination of $\tilde W_0$'s elements inside of each Hermite polynomial.

 We use the properties of Hermite polynomials \citep{olver2010nist}[\S 18.18.11]:
\begin{align}\small
\label{eq:181810}
\frac{(a_{1}^{2}+\cdots+a_{r}^{2})^{\frac{i}{2}}}{i!}\tilde q_{i}\Big(\frac{a_1 x_1+\cdots+a_r x_r}{(a_{1}^{2}+\cdots+a_{r}^{2})^{\frac{1}{2}}}\Big)=\!\!\!\!\!\!\!\sum_{s_{1}+\cdots+s_{r}=i}\frac{a_{1}^{s_{1}}\cdots a_{r}^{s_{r}}}{s_{1}!\cdots s_{r}!}\tilde q_{s_{1}}(x_{1})\cdots \tilde q_{s_{r}}(x_{r})
\end{align}
where $\tilde q_i$'s form an orthogonal basis, equipped with the inner product  $\langle u, v \rangle_{\tilde\Hc} =\frac{1}{\sqrt{\pi}} \int u(x) v(x) \exp(-x^2)\der x$.  This basis follows the physicist's convention of Hermite polynomial.

Since $\tilde q_i$ and $q_i$ are rescalings of the other, we can replace $q_i$'s into ~\eqref{eq:181810}. Note that we have $\|\xbf_a\|_2=1$ for all $a\in[n]$. Then we have
\begin{align}\label{eq:dlmf-prop}
q_i(\xbf_a^\top \tilde W_{0,c,\ra} ) & = i! \sum_{s_1+\cdots+s_{d_0} = i} \frac{x_{a,1}^{s_1}\cdots x_{a,d_0}^{s_{d_0}}}{s_1!\cdots s_{d_0}!} q_{s_1}(\tilde W_{0,c,1})\cdots q_{s_{d_0}}(\tilde W_{0,c,d_0}) 
\end{align}
where $x_{a,k}$ and $\tilde W_{0,c,k}$ are $k$-th entry of $\xbf_a$ and $\tilde W_{0,c,\ra}$ for $k\in[d_0]$.  Using the expansion in  \eqref{eq:dlmf-prop}, we expand  \eqref{eq:entry} as follows: 
\begin{align}\label{eq:summand-for-entries-of-outerprod-Ex2}
\begin{split}
\z_{i,j}(a,b)
& =i! j!\sum_{s_1+\cdots + s_{d_0}=i}\sum_{s'_1+\cdots+s'_{d_0}=j} 
\frac{x_{a,1}^{s_1}\cdots x_{a,d_0}^{s_{d_0}}}{s_1!\cdots s_{d_0}!} 
\cdot \frac{x_{b,1}^{s'_1}\cdots x_{b,d_0}^{s'_{d_0}}}{s'_1!\cdots s'_{d_0}!} \rho_{\sbf,\sbf'}(\tilde W_{0,c,\ra})\\
& = 
\begin{cases}
( i!)^2 \sum_{s_1+\cdots +s_{d_0} = i} \frac{(x_{a,1}x_{b,1})^{s_1} \cdots (x_{a,d_0} x_{b,d_0})^{s_{d_0}} }{s_1!\cdots s_{d_0}!} & i=j, \\
0 & i\ne j
\end{cases}\\
& = 
\begin{cases}
i! \sum_{s_1+\cdots +s_{d_0} = i}{i \choose s_1,\cdots,s_{d_0}} (x_{a,1}x_{b,1})^{s_1} \cdots (x_{a,d_0} x_{b,d_0})^{s_{d_0}} & i=j, \\
0 & i\ne j
\end{cases}
\end{split}
\end{align} where $\z_{i,j}(a,b)=\E \l[ q_i(\xbf_a^\top \tilde W_{0,c,\ra} )q_j(\tilde W_{0,c,\ra}^{\top} \xbf_b ) \r]$, 
\begin{align}\nn
\rho_{\sbf,\sbf'}(\tilde W_{0,c,\ra})=\E \l[ 
q_{s_1}(\tilde W_{0,c,1})\cdots q_{s_{d_0}}(\tilde W_{0,c,d_0}) 
\cdot q_{s'_1}(\tilde W_{0,c,1})\cdots q_{s'_{d_0}}(\tilde W_{0,c,d_0}) 
\r],
\end{align} $\sbf=[s_1,\cdots,s_{d_0}]$, and $\sbf'=[s'_1,\cdots,s'_{d_0}]$.

%
To simplify the expression in~\eqref{eq:summand-for-entries-of-outerprod-Ex2},  we define $X^{*i} \in \R^{ d_0^i \times n}$ where the $a$-th column is given by 
\begin{align}\nn
X^{*i}_{a} = \vector(\xbf_a \otimes \cdots \otimes \xbf_a) \in \R^{ d_0^i},
\end{align}
which is also called Khatri-Rao product. For $i = 0$, we use the convention that $X^{*0} =\mathbf{1}\mathbf{1}^\top\in\R^{n\times n}$. 

We can rewrite \eqref{eq:summand-for-entries-of-outerprod-Ex2} as follows:  
\begin{align}\label{eq:entries-final-Ex2}
\z_{i,j}(a,b) & = 
\begin{cases}
i! \langle X^{*i}_{a} , X^{*i}_{b} \rangle & i = j \\
0 & i\ne j.
\end{cases} 
\end{align}

Substituting \eqref{eq:entries-final-Ex2} back into \eqref{eq:entry}, we find that 
\begin{align}
\label{eq:entries-prefinal-Ex2-nocompact}
\begin{split}
\l[ \E[ q_i(X^\top \tilde W_0^\top) q_j(\tilde W_0 X)] \r]_{a,b}
& = \sum_{c=1}^{d_1} \E \l[ q_i(\xbf_a^\top \tilde W_{0,c,\ra} )q_j(\tilde W_{0,c,\ra}^{\top} \xbf_b ) \r]\\
& = \begin{cases}
d_1 i! \langle X^{*i}_{a} , X^{*i}_{b} \rangle & i = j  \\
0 & i\ne j.
\end{cases}
\end{split}
\end{align} 

 Substituting \eqref{eq:entries-prefinal-Ex2-nocompact} into \eqref{eq:prod-of-decomps-Ex2}, we have 
\begin{align}\label{eq:exp-M0-full}
\E\l[\tilde M_0 \r]  =   d_1 \l(c_0^2 \mathbf{1}\mathbf{1}^\top + c_1^2  X^\top X +  \sum_{i=2}^\infty \frac{c_i^2}{i!}   (X^{*i})^{\top} X^{*i}\r).            
\end{align}

We now establish an upper  bound on $\s_{\m}\l(\sum_{i=2}^\infty \frac{c_i^2}{i!}   (X^{*i})^{\top} X^{*i}\r)$:
\begin{align}\label{eq:observe-that}
\begin{split}
\s_{\m}\l( \sum_{i=2}^\infty \frac{c_i^2}{i!}  (X^{*i})^{\top} X^{*i}  \r) & \le \sum_{i=2}^{\infty} \frac{c_i^2}{i!} \s_{\max}( (X^{*i})^{\top} X^{*i} ) \\
& \leq c_{\infty}^2 \s^2_{\max}(X)
\end{split} 
\end{align}
where $c_{\infty}$ is given by  
\begin{align}\nn
c_{\infty}^2 = \sum_{i=2}^{\infty} \frac{c_i^2}{i!},
\end{align} 
which is finite provided that $\|\phi\|_{\Hc}$ is bounded. 

Using~\eqref{eq:observe-that}, we now establish an upper bound on $\s_{\m}(\E[\tilde M_0])$:  
\begin{align}\nn
\s_{\m}( \E[\tilde M_0] )\lesssim d_1\l(n c_0^2  + (c_1^2+ c_{\infty}^2)  \s^2_{\max}(X) \r).  
\end{align}

Moreover, suppose there exists some $t$ such that $\s_{\min}(X^{*t}) > 0$.  This requires to have $d_0^t \ge n$. 
Putting together the lower bound on $\s_{\min}( \E[\tilde M_0] )$ and the upper bound on $\s_{\min}( \E[\tilde M_0] )$, noting $W_0=\o_1\tilde W_0$ and applying $\tau^{r_1}\phi(a) \leq \phi(\tau a) \leq\tau^{r_2}\phi(a)$, we have
\begin{align}\label{eq:upp-low-bnd-on-EM0}
\o_1^{2r_1}d_1 \frac{c_t^2}{t!}\s^2_{\min}(X^{*t}) \lesssim \s_{\min}( \E[M_0] ) \le \s_{\m}( \E[M_0] ) \lesssim \o_1^{2r_2}d_1\l(n c_0^2  + (c_1^2+ c_{\infty}^2)  \s^2_{\max}(X) \r).
\end{align}

\subsection{Concentration of the random matrix $M_0$} 

To see how well the random matrix $M_0$  concentrates about its expectation, note that 
\begin{align}\label{eq:defn-A-Ex2}
\begin{split}
M_0 & =  \phi(X^\top W_0^{\top}) \phi (W_0 X) \\
& = \sum_{i=1}^{d_1} \phi(X^\top W_{0,i,\ra}^{\top}) \phi( W_{0,i,\ra} X )\\
& =\sum_{i=1}^{d_1} A_i
\end{split}
\end{align}
where $\{A_i\}_{i=1}^{d_1}\subset \R^{n\times n}$ are independent random matrices. 

\noindent Consider the event $\Ec_1$ that 
\begin{align}\label{eq:event}
\max_{i\in [d_1]} \|W_{0,i,\ra}\|_2 \lesssim  k_1\o_1\sqrt{d_0\log d_1},
\quad 
\max_{i\in [d_1]} \|V_{0,i,\da}\|_2 \lesssim k_2\o_2 \sqrt{d_2\log d_1}
\end{align}
where $V_{0,i,\da}$ is the $i$-th column of $V_0$. 
Note that $W_{0,i,\ra}\in \R^{d_0}$ and $V_{0,i,\da}\in \R^{d_2}$ are  random zero-mean Gaussian vectors whose entries' variances are $\o_1^2 $ and $\o_2^2$, respectively. Therefore, with an application of the scalar Bernstein inequality \citep[Proposition 5.16]{vershynin2010introduction}, followed by the union bound, we observe that  the event $\Ec_1$ happens except with a probability of at most

\begin{align}\label{eq:defn-pr1}
p_1 :=  d_1^{-Ck_1 d_0}+d_1^{-Ck_2 d_2} ,
\end{align}
for a universal constant $C$ with sufficiently large $k_1, k_2$. 

Let $i\in[d_1]. $ Conditioned on the event $\Ec_1$, an upper bound on $\| \phi(X^\top W_{0,i,\ra}) \|_2$ is given by:  

\begin{align}\label{eq:pre-bnd-A-Ex2}
\| \phi(X^\top W_{0,i,\ra}) \|_2 & \lesssim  \dot{\phi}_{\m} \s_{\m}(X) k_1\o_1\sqrt{d_0 \log d_1}.
\end{align}

Moreover, we have 
\begin{align}\label{eq:max-bnd-for-chernoff}
\begin{split}
\s_{\m}(A_i) & = \| \phi( X^\top W_{0,i,\ra}) \|_2 ^2\\ 
& = \| \phi(X^\top W_{0,i,\ra}) - \phi(0) \|_2^2 \\
& \lesssim  \dot{\phi}_{\m}^{2} \s^2_{\m}(X) k_1^2\o_1^2 d_0 \log d_1.
\end{split}
\end{align}

We now focus on the concentration of $\s_{\min} (M_0)$ and $\s_{\m} (M_0)$. We use a concentration property, which provides the tail bound of $\tilde f(W) = \phi(X^{\top}W^{\top})\phi(WX)$ with multivariate Gaussian input $W$. In the following lemma, we show that $\tilde f$ is  a Lipschitz function, and its Lipschitz constant explains how $\tilde f(W)$ concentrates around its mean.

\blm	\label{lemma:lipschitz-const-f}
	Let $\tilde f(W) = \phi(X^{\top}W^{\top})\phi(WX)$. Suppose $W$ satisfies~\eqref{eq:event}. Then $\tilde f$ is $\kappa$-Lipschitz function with constant $\kappa = 4\dot{\phi}_{\m}^2 \s^2_{\m}(X) k_1\o_1\sqrt{d_0 \log d_1}$. So we have
	\begin{align}\nn
	\|\tilde f(W) - \tilde f(W')\| < 4\dot{\phi}_{\m}^2  \s^2_{\m}(X) k_1\o_1\sqrt{d_0\log d_1} \cdot  \|W-W'\|.
	\end{align}
\elm
\bpr
Note that $\tilde f(W_0) = M_0$ and $\tilde f$ can be represented as
\begin{align}\nn
\tilde f(X) = \sum_{i=1}^{d_1} f_i(W_{i, \ra})
\end{align}
where $f_i$ is given by $f_i(W_{i, \ra}) = \phi(X^\top W^{\top}_{i,\ra}) \phi( W_{i,\ra} X )$. We prove that each $f_i$ is $\kappa$-Lipschitz, which implies that $\tilde f$ is also $\kappa$-Lipschitz. 

We note that $f_i$'s can be expressed as a composition of three functions: 
\begin{align}\nn
f_i(\vbf) = (g_{1}\circ g_{2} \circ g_3)(\vbf)
\end{align}
where $g_1, ~g_2$, and $g_3$ are given by 
\begin{align}
g_1(\vbf) = \vbf\vbf^{\top}, \ f_2(\vbf) = \phi(\vbf), \ f_3(\vbf) = \vbf X.
\end{align}

It is clear that $g_2$ is $\dot{\phi}_{\m}$-Lipschitz, and $g_3$ is $\s_{\m}(X)$-Lipschitz from their definitions. Lipschitz constant of $g_1$ comes from the domain bound as follows: 
\begin{align}\label{eq:lipschitz-f1]}
\begin{split}
\|g_1(\vbf+\delta \vbf) - g_1(\vbf)\| &= \|\delta{\vbf}\vbf^{\top} + \vbf \delta \vbf^{\top} + \delta {\vbf} \delta \vbf^{\top}\| \\
& \leq 2\|\delta \vbf{\vbf}^{\top}\| + \|\delta {\vbf} \delta\vbf^{\top}\|\\
& \leq (2\|\vbf\| + \|\delta \vbf\|) \cdot \|\delta \vbf\|.
\end{split}
\end{align}

A bound on $(2\|\vbf\| + \|\delta \vbf\|)$ is obtained in \eqref{eq:pre-bnd-A-Ex2}. Then $g_1$ is $\kappa_1$-Lipschitz function with $\kappa_1 = 4\dot{\phi}_{\m} \s_{\m}(X) k_1\o_1\sqrt{d_0 \log d_1}$. Therefore, all $g_1, ~g_2$ and $g_3$ are Lipschitz function, so their composition $f_i$ is also Lipschitz function with constant $\kappa = 4\dot{\phi}^2_{\m} \s^2_{\m}(X) k_1\o_1\sqrt{d_0 \log d_1}$, which completes the proof.  
\epr

\blm\label{lm:MultiGaussian}
Let $\zbf\in\R^d$ denote a Gaussian random vector. Then we have 
$\Pr\{ \|\zbf - \E[\zbf]\| > t \ | \Ec_2\} \lesssim \exp(-t^2)$ where $\Ec_2$ is the event that $\|\zbf\|$ is bounded. 
\elm
We can focus on the tail distribution of $M_0 = \tilde f(W_0)$. Using Lemmas~\ref{lemma:lipschitz-const-f} and \ref{lm:MultiGaussian}, we have
\begin{align}\label{eq:gaussian-concentration} 
\Pr\{\|M_0 - \E [M_0]\| > t  \ | \Ec_1 \} \lesssim \exp(-k_3^2)
\end{align}
where $t = k_3 4\dot{\phi}^2_{\m} \s^2_{\m}(X) k_1\o_1\sqrt{d_0 \log d_1}$ with some constant $k_3$. 

Using \eqref{eq:gaussian-concentration}, we now establish a tail bound on $\s_{\min} (M_0)$: 
\begin{align}\nn
\begin{split}
\Pr\{ \s_{\min} (M_0) \le (1-\delta_1) \s_{\min}(\E [M_0]) | \Ec_1 \} & \le \Pr\{ |\s_{\min} (M_0) -  \s_{\min}(\E [M_0])| \ge \delta_1 \s_{\min}(\E [M_0]) | \Ec_1 \} \\
& \le \Pr\{ \s_{\min} (M_0-\E [M_0]) \ge \delta_1 \s_{\min}(\E [M_0]) | \Ec_1 \} \\
& \le \Pr\{ \s_{\m} (M_0-\E [M_0]) \ge \delta_1 \s_{\min}(\E [M_0]) | \Ec_1 \} \\
& \le \Pr\{ \|M_0 -  \E[M_0]\| \ge \delta_1 \s_{\min}(\E [M_0])  | \Ec_1 \}  \\
& \lesssim p_2 
\end{split}
\end{align}
where 
\begin{align}\nn
p_2 = \exp\l(-\l(\frac{\delta_1 \s_{\min}(\E [M_0])}{4\dot{\phi}^2_{\m} \s^2_{\m}(X) k_1\o_1\sqrt{d_0 \log d_1}} \r)^2\r).
\end{align}

Similarly, we obtain
\begin{align}\nn
\Pr\{ \s_{\max} (M_0) \geq (1+\delta_2) \s_{\max}(\E [M_0]) | \Ec_1 \}\lesssim p_3
\end{align} where 
\begin{align}\nn
p_3 = \exp\l(-\l(\frac{\delta_2 \s_{\m}(\E [M_0])}{4\dot{\phi}^2_{\m} \s^2_{\m}(X) k_1\o_1\sqrt{d_0 \log d_1}} \r)^2\r).
\end{align}

Putting these bounds together with~\eqref{eq:upp-low-bnd-on-EM0}, we have :
\begin{align}\label{eq:estimates-for-smin-smax}
\begin{split}
\o_1^{r_1}\sqrt{(1-\delta_1)  \frac{c_t^2}{t!} d_1 } \s_{\min}(X^{*t}) &\leq \s_{\min}( \phi(W_0 X) )\\
 \s_{\m}( \phi(W_0 X) ) &\leq \sqrt{(1+\delta_2)} \o_1^{r_2}(\sqrt{(c_1^2+c_\infty^2) d_1} \s_{\m}(X) + |c_0|\sqrt{d_1n})
\end{split}
\end{align} except with a probability of at most $p_1+p_2+p_3$. 

With establishing the bounds on $\s_{\min}( \phi(W_0 X))$ and $\s_{\m}( \phi(W_0 X))$, we can finally estimate $\mu_{\Phi}, \nu_{\Phi}$ as follows:  

\subsection{Lower bound on $\mu_{\Phi}$}
A lower bound on $\mu_{\Phi}$ is given by
\begin{align}
\o_1^{r_1}\sqrt{(1-\delta_1) \frac{c_t^2}{t!} d_1 } \s_{\min}(X^{*t})  \leq \s_{\min}( \phi(W_0 X) ) = \mu_{\Phi},
\end{align}

except with a probability of at most $p_1+p_2$.

\subsection{Upper bound on $\nu_{\Phi}$}\label{sec:UBnu}
Since $\nu_\Phi  = \dot{\phi}_{\m} \s_{\m}(X) \s_{\m}(V_0)+\s_{\max}(\phi( W_0 X)) $, we obtain a bound on $\s_{\max}(V_0)$:  

Since $V_0$ is a Gaussian random matrix, we have  
\begin{align}\label{eq:rand-gauss-1}
\s_{\max}(V_0) \le \o_2(2\sqrt{d_1}+\sqrt{d_2} ) \lesssim \o_2 \sqrt{d_1}
\end{align}
except with a probability of at most $p_4 = \exp(-C d_1) $ where $C$ is a universal constant~\citep{vershynin2010introduction}[Corollary 5.35].  

Combining ~\eqref{eq:rand-gauss-1} with the upper bound on $\s_{\m}( \phi(W_0 X))$,  we have
\begin{align}\nn
\begin{split}
\nu_\Phi & =  \dot{\phi}_{\m} \s_{\m}(X) \s_{\m}(V_0)+\s_{\max}(\phi( W_0 X)) \\
& \lesssim  \o_2 \dot{\phi}_{\m}\s_{\max}(X) \sqrt{d_1}+\o_1^{r_2}\sqrt{(1+\delta_2)(c_1^2+c_{\infty}^2) d_1} \s_{\max}(X) + \o_1^{r_2}|c_0|\sqrt{(1+\delta_2)d_1n} 
\end{split} 
\end{align}
except with a probability of at most $p_1+p_3+p_4$.

\subsection{Upper bound on $h(\Theta_0)$}
In this section, we bound $h(\Theta_0)$.  Using $\|\abf+\bbf\|_2^2 \le 2\|\abf\|_2^2 +2\|\bbf\|_2^2$, we have    
\begin{align}\label{eq:bnd-h-Ex2}
\begin{split}
h(\Theta_0) &= \| V_0 \phi( W_0 X) - Y\|^2 \\
& \le 2\| V_0 \phi( W_{0} X) \|^2 +2 \|Y\|^2.
\end{split}
\end{align}
To upper bound the random norm in~\eqref{eq:bnd-h-Ex2}, we first decompose $V_0 \phi( W_{0} X)$ into terms including $W_{0,i,\ra}\in \R^{d_0}$ and $V_{0,i,\da}\in \R^{d_2}$ as follows: 
\begin{align}\label{eq:defn-Bis-ex2}
V_0 \phi (W_{0} X)   =  \sum_{i=1}^{d_1} B_i
\end{align}
where $B_i= V_{0,i,\da} \phi( W_{0,i,\ra}^{\top} X)\in \R^{d_2 \times n}$'s are independent random matrices for $i\in[d_1]$.

Conditioned on the event $\mathcal{E}_1$ defined in \eqref{eq:event}, we bound $\|B_i\|$:  
\begin{align}\label{eq:bnd-on-B-ex2}
\begin{split}
\|B_i\| & = \| V_{0,i,\da}\|_2 \| \phi( W_{0,i,\ra}^{\top} X) \|_2 \\
& \leq \|V_{0,i,\da}\|_2 \cdot  \dot{\phi}_{\m}  \s_{\m}(X) k_1\o_1 \sqrt{d_0 \log d_1}\\
& \leq \o_1\o_2 \dot{\phi}_{\m}  \s_{\m}(X) k_1 k_2 \sqrt{d_0d_2} \log d_1
\end{split}
\end{align} 
for $i\le d_1$. 

Substituting the upper bound in \ref{eq:defn-Bis-ex2} into \ref{eq:bnd-on-B-ex2} and applying  the  Hoeffding inequality~\citep{Hoeffding}, we have  
\begin{align}\nn
\begin{split}
\Pr\{ \| V_0 \phi(W_{0}X) \| \gtrsim  u(d_0,d_1,d_2)| \Ec_1 \}
& = \Pr\{ \| V_0 \phi( W_{0}X) - \E[V_0 \phi( W_{0}X))| \Ec_1] \| \gtrsim u(d_0,d_1,d_2) | \mathcal{E}_1 \} \\
&\leq \Pr\l\{\sum_{i=1}^{d_1} \|B_i-\E[B_i]\| \gtrsim u(d_0,d_1,d_2) | \mathcal{E}_1 \r\}\\
& \le p_5
\end{split}
\end{align}
where 
\begin{align}\nn
u(d_0,d_1,d_2)=\delta_3\o_1 \o_2 \dot{\phi}_{\m} k_1 k_2 \sqrt{d_0d_1d_2} \s_{\m}(X)  \log d_1
\end{align} and $p_5=\exp(-C\delta_3^2)$
 with $\delta_3\ge 0$ and a universal constant $C$.

Therefore, under the event $\Ec_1$, we have 
\begin{align}\label{eq:bnd-h-ex2}
\begin{split}
h(\Theta_0) & \le 2\| V_0 \phi( W_0 X) \|^2 + 2\|Y\|^2  \\
& \lesssim \delta_3^2 \o_1^2\o_2^2 \dot{\phi}^2_{\m} k_1^2 k_2^2 d_0 d_1 d_2  \s^2_{\m}(X)\log^2 d_1 + \|Y\|^2
\end{split}
\end{align}
except with a probability of at most $p_1+p_5$.  It is natural to assume that $d_2=o(d_1)$. We also have $\|Y\| \leq 1$. 
 
Suppose that 
\begin{align}\label{eq:assump-o2-ex2}
\o_1\o_2 \lesssim \frac{1}{ \dot{\phi}_{\m}\sqrt{d_0 d_1} \log d_1}.
\end{align}

Substituting \eqref{eq:assump-o2-ex2} into \eqref{eq:bnd-h-ex2}, we have 
\begin{align}\label{eq:bnd-h-ex2-final}
h(\Theta_0) & \le \delta_3^2 k_1^2 k_2^2 \s^2_{\m}(X)
\end{align}
where $\delta_3, ~k_1$, and $k_2$ are all constants and independent of $d_0, ~d_1$,  and $n$.

\subsection{Denouement \label{sec:denoument}}

The key condition for linear rate convergence of gradient descent in ~\eqref{eq:key-assump-equiv} is 
\begin{align}\nn
h(\Theta_0) \lesssim  \frac{\alpha_f \mu_{\Phi}^6}{\beta_{\Phi}^2 \nu_{\Phi}^2}.
\end{align}

Putting everything together for the shallow neural network, with high probably,  we have 
\begin{align}\label{eq:den}
\begin{split}
\alpha_f & = 2 \\
 \nu_{\Phi} & = \o_2 \dot{\phi}_{\m}\s_{\max}(X) \sqrt{d_1}+\sqrt{(1+\delta_2)\o_1^{2r_2}(c_1^2+c_\infty^2)} \s_{\m}(X) \sqrt{d_1} + |c_0|\sqrt{\o_1^{2r_2}(1+\delta_2)d_1n} \\
\mu_{\Phi} & = \o_1^{r_1}\sqrt{(1-\delta_1) \frac{c_t^2}{t!} d_1 } \s_{\min}(X^{*t})\\
\beta_{\Phi} & = \sqrt{2}  \s_{\m}(X) \l( \dot{\phi}_{\m} + \ddot{\phi}_{\m} \chi_{\m} \r). 
\end{split}
\end{align}
We note that the order of $\s_{\m}(X)$ and $\s_{\min}(X^{*t})$ play significant roles for the \op order analysis. For $t=1$, it requires $n \simeq d_0$, which is not a common setting in practice.  In the following, we focus on $t \ge 2$.

\subsection{Order analysis with $t\ge 2$} 
In this section, we assume $|c_0|$ is sufficiently large such that $|c_0|\sqrt{(1+ \delta_2)d_1n}$ becomes the dominating term in $\nu_{\Phi}$.\footnote{To have a nonzero $c_0$, the activation function should not be an odd function.} Then a sufficient condition to satisfy  \eqref{eq:key-assump-equiv} is

\begin{align}\label{eq:second-overparam-bnd-ex22}
d_1^2 \gtrsim
\frac{  \delta_3^2 c_0^2(1+\delta_2) k_1^2 k_2^2 (\dot{\phi}_{\m}+\ddot{\phi}_{\m}\chi_{\m})^2    \s^4_{\max}(X) n t!^3}{\o_1^{6r_1-2r_2}(1-\delta_1)^3 c_t^6 \s^6_{\min}(X^{*t})},
\end{align}
which can be written as 
\begin{align}\nn
d_1 & \gtrsim \sqrt{\frac{\delta_3^2 c_0^2(1+\delta_2) k_1^2 k_2^2 (\dot{\phi}_{\m}+\ddot{\phi}_{\m}\chi_{\max})^2t!^3}{\o_1^{6r_1-2r_2}(1-\delta_1)^3c_t^6}} \cdot \frac{\sqrt{n} \s^2_{\m}(X)}{\s^3_{\min}(X^{*t})}.
\end{align}
For notational simplicity, we let $\delta_4 =\max( k_1, k_2)$ and denote $\Cc_\delta = \{\delta_1, \delta_2, \delta_3, \delta_4\}$ and 
\begin{align}
\xi(\Cc_\delta, t, \phi, \{c_i\}_{i\geq 0}) = \sqrt{\frac{\delta_3^2 c_0^2(1+\delta_2)\delta_4^4 (\dot{\phi}_{\m}+\ddot{\phi}_{\m}\chi_{\m})^2t!^3}{\o_1^{6r_1-2r_2}(1-\delta_1)^3c_t^6}}.
\end{align}
Note that $\xi(\Cc_\delta, t, \phi, \{c_i\}_{i\geq 0})$ can be viewed as a constant w.r.t. $d_0, ~d_1$, and $n$. Then~\eqref{eq:second-overparam-bnd-ex22} can be written as: 
\begin{align}
d_1 = \tilde{\Omega}(\frac{\sqrt{n} \s^2_{\max}(X)}{\s^3_{\min}(X^{*t})}).
\end{align}
It remains to estimate $\s_{\m}(X)$ and $\s_{\min}(X^{*t})$  to finish the order analysis of $d_1$. Suppose that $n \simeq d_0^t$. Then , along the lines of \citep{oymak2019towards}[Section 2.1], we have $\s_{\m}(X) \simeq \sqrt{\frac{n}{d_0}}$ and $\s_{\min}(X^{*t}) \simeq \sqrt{\frac{n}{d_0^t}} \simeq 1$. 

Combining them all, we have
\begin{align}
d_1 & \gtrsim \xi(\Cc_\delta, t, \phi, \{c_i\}_{i\geq 0})  \frac{n^{\frac{3}{2}}}{d_0}.
\end{align}

Therefore, the overall \op degree becomes $d_0d_1 \simeq \tilde{\Omega}(n^{\frac{3}{2}})$ for $t \ge 2$.


The exact expression of $\psi(\phi,\xi, ,d_0,d_1,d_2,X)$ in Theorem \ref{thm:shallow-net-ex2} is given by
\begin{align}
\psi & \leq p_1 + p_2 + p_3 + p_4 + p_5 \nn \\
& \leq d_1^{-C \delta_4 d_0}+d_1^{-C \delta_4 d_2} + e^{-\l(\frac{\delta_1 \s_{\min}(\E[M_0])}{4\dot{\phi}_{\m}^2 \s^2_{\m}(X) \delta_4\sqrt{d_0 \log d_1}} \r)^2} + e^{-\l(\frac{\delta_2 \s_{\max}(\E[M_0])}{4\dot{\phi}_{\m}^2 \s^2_{\m}(X) \delta_4\sqrt{d_0 \log d_1}} \r)^2} + e^{-C d_1} + e^{-C\delta_3^2}. \nn 
\end{align}
Note that $d_1^{-C \delta_4 d_0}+d_1^{-C \delta_4 d_2} +  \exp(-C d_1) + \exp(-C\delta_3^2)$ decreases exponentially, which can be sufficiently small without changing the order of $d_1$. 


Finally,  with $d_0d_1 \simeq \tilde{\Omega}(n^{\frac{3}{2}})$, the gradient descent converges to a global minimum with linear rate with probability at least $1- \psi$, which can be arbitrary small.

\paragraph{Order analysis without boundedness assumption on $\s_{\max}( V_k)$ in Assumption~\ref{assumption2}. }

So far, we assumed $\s_{\max}( V_k)$ is bounded for $k\geq 0$.  We can relax this assumption by bounding the length of the trajectory of gradient descent as discussed in Appendix~\ref{app:thm:gd-final-result}.  Recall \eqref{eq:trabound}: 

\begin{align}\nn
\ell(I) \lesssim \frac{\nu_\Phi \sqrt{f(Z_0)}  }{ \sqrt{\a_f} \mu_\Phi^2}.
\end{align}

Using triangular inequality and substituting~\eqref{eq:trabound},  we can obtain a bound on $\|V_k\|$
\begin{align}\label{eq:chibound}
\begin{split}
\|V_k\| &\leq \|V_k - V_0\| + \|V_0\| \\
&\leq \frac{\nu_\Phi \sqrt{f(Z_0)}  }{ \sqrt{\a_f} \mu_\Phi^2} + \|V_0\|
\end{split}
\end{align}

As shown in~\eqref{eq:rand-gauss-1}, $\|V_0\|\lesssim \o_2 \sqrt{d_1}$ with high probability over the choice of $V_0$.  With sufficiently small $ \o_2$, the first term in the upper bound dominates in \eqref{eq:chibound}.  Applying ~\eqref{eq:bnd-h-ex2-final} and substituting~\eqref{eq:chibound} into \eqref{eq:second-overparam-bnd-ex22}, we have 
\begin{align}\nn
d_1^3 &\gtrsim \frac{n^2\s^6_{\max}(X)}{\s^{10}_{\min}(X^{*t})}  \\
d_1 &\gtrsim \frac{n^{\frac{5}{3}}}{d_0}.\nn
\end{align}

The overall \op degree becomes $d_0d_1 \simeq \tilde{\Omega}(n^{\frac{5}{3}})$, which is slightly worse than the result of Theorem \ref{thm:shallow-net-ex2} under boundedness assumption on $\s_{\max}( V_k)$. Note that we still have a subquadratic scaling on the network width.

\section{Additional discussion on lazy training in Section~\ref{sec:experiment}}\label{app:lazy}
In this section, we provide an asymptotic analysis for the term $\|h(\Theta_i) - \tilde{h}(\tilde{\Theta}_i)\|$ to show that there exists a regime where our initialization can avoid lazy training. Recall our setting:
\begin{align}\nn
\Phi(\Theta) = V \cdot \phi(WX)
\end{align}
where $W \sim \Nc(0, \o_1^2)$ and $V \sim \Nc(0, \o_2^2)$. Following the theoretical guidance in \eqref{eq:init-ex2-2}, we set $\o_1 \o_2 \simeq \frac{1}{\sqrt{d_0d_1}}$. 

An upper bound on $\|h(\Theta_i) - \tilde{h}(\tilde{\Theta}_i)\|$ is given by \citep[Theorem 2.3]{chizat2019lazy}:
\begin{align}\label{eq:lazy_appendix}
\|h(\Theta_i) - \tilde{h}(\tilde{\Theta}_i)\| \lesssim \frac{\text{Lip}(\Der\Phi(\Theta))}{\text{Lip}(\Phi(\Theta))^2}.
\end{align}

In the following, we estimate $\frac{\text{Lip}(\Der\Phi(\Theta))}{\text{Lip}(\Phi(\Theta))^2}$ to find when it is not bound to be close to zero.

Substituting $\beta_{\Phi}$ and $\nu_{\Phi}$ expressions in \eqref{eq:den} into the upper bound in~\eqref{eq:lazy_appendix} for sufficiently large $n,c_0$, we have  
\begin{align}
\|h(\Theta_i) - \tilde{h}(\tilde{\Theta}_i)\| & \lesssim
\frac{\sqrt{2}\s_{\m}(X) ( \dot{\phi}_{\m}+ \ddot{\phi}_{\m} \chi_{\m})}{(\o_2 \dot{\phi}_{\m}\s_{\m}(X)\sqrt{d_1} + \o_1^{r_2}c_0 \sqrt{(1+\delta_2)d_1n})^2}.
\end{align}

We now find an upper bound on $\chi_{\m}$ by bounding the total length of the trajectory of gradient descent as in Appendix ~\ref{app:thm:gd-final-result} where the length of the trajectory traced by gradient descent is given by \eqref{eq:trabound}:  
\begin{align}\nn
\ell(I) \le \frac{\nu_\Phi \sqrt{f(Z_0)}  }{ \sqrt{\a_f} \mu_\Phi^2}.
\end{align}

Using \eqref{eq:trabound}, \eqref{eq:rand-gauss-1}, and \eqref{eq:bnd-h-ex2-final}, a bound on $\chi_{\m}$ is given by 

\begin{align}
\begin{split}
\|V_i\|_2 &\leq \|V_i - V_0\|_F + \|V_0\|_2\\
&\leq \frac{\nu_\Phi \sqrt{f(Z_0)}  }{ \sqrt{\a_f} \mu_\Phi^2} + \|V_0\|_2  \\
&\lesssim \frac{(\o_2\dot{\phi}_{\m} \s_{\m}(X) + \o_1^{r_2} c_0\sqrt{n})\s_{\m}(X)}{\o_1^{2r_1}\sqrt{d_1}\s_{\min}^2(X^{*t})} + \o_2\sqrt{d_1}
\end{split}
\end{align}
Therefore we have 
\begin{align}\nn
\|h(\Theta_i) - \tilde{h}(\tilde{\Theta}_i)\| & \lesssim \frac{\sqrt{2}\s_{\max}(X) \l( \dot{\phi}_{\m} +  \ddot{\phi}_{\m}\frac{(\o_2\dot{\phi}_{\m} \s_{\m}(X) + \o_1^{r_2} c_0\sqrt{n})\s_{\m}(X)}{\o_1^{2r_1}\sqrt{d_1}\s_{\min}^2(X^{*t})} + \o_2\ddot{\phi}_{\m}\sqrt{d_1}\r)}{(\o_2 \dot{\phi}_{\m}\s_{\m}(X)\sqrt{d_1} + \o_1^{r_2}c_0 \sqrt{(1+\delta_2)d_1n})^2}
\end{align}

We now consider two cases: 1) $ \o_2\dot{\phi}_{\m} \s_{\max}(X) \gtrsim \o_1^{r_2} c_0\sqrt{n}$ and 2) $ \o_2\dot{\phi}_{\m} \s_{\max}(X) \lesssim \o_1^{r_2} c_0\sqrt{n}$. More precisely, for the asymptomatic analysis, we consider extremal cases $\o_1 \gg \o_2$ and $\o_1 \ll \o_2$ and evaluate $\|h(\Theta_i) - \tilde{h}(\tilde{\Theta}_i)\|$ in each case: 

\subsection{Regime with $\o_2 \gg \o_1$ }

In the \op regime with large $d$,  we note that $\ddot{\phi}_{\m}\frac{(\o_2\dot{\phi}_{\m} \s_{\m}(X) + \o_1^{r_2} c_0\sqrt{n})\s_{\m}(X)}{\o_1^{2r_1}\sqrt{d_1}\s_{\min}^2(X^{*t})} + \o_2\ddot{\phi}_{\m}\sqrt{d_1}\gtrsim  \dot{\phi}_{\m}$. Then we have

\begin{align}\nn
\begin{split}
\|h(\Theta_i) - \tilde{h}(\tilde{\Theta}_i)\| & \lesssim \frac{\sqrt{2}\s_{\max}(X) \l(\frac{(\o_2\dot{\phi}_{\m} \s_{\m}(X) + \o_1^{r_2} c_0\sqrt{n})\s_{\m}(X)}{\o_1^{2r_1}\sqrt{d_1}\s_{\min}^2(X^{*t})} + \o_2\sqrt{d_1}\r)}{(\o_2 \dot{\phi}_{\m}\s_{\m}(X)\sqrt{d_1} + \o_1^{r_2}c_0 \sqrt{(1+\delta_2)d_1n})^2}\\
& \lesssim \frac{\s_{\max}^2(X) \left(\frac{\o_2}{\o_1^{2r_1}\sqrt{d_1}\s_{\min}^2(X^{*t})}\right)}{(\o_2 \s_{\m}(X) + \o_1^{r_2} c_0\sqrt{n})^2 d_1}  \\
& \lesssim \frac{\s_{\max}^2(X) \o_2/ d_1^{\frac{3}{2}}}{\s_{\min}^2(X^{*t})(\o_1^{r_1}\o_2 \s_{\m}(X) + \o_1^{r_1+r_2} c_0\sqrt{n})^2}\\
& \lesssim \frac{\s_{\max}^2(X) \o_2/ d_1^{\frac{3}{2}}}{\l(\s_{\min}(X^{*t}) \s_{\m}(X) \frac{ \o_1^{r_1-1}}{\sqrt{d_0d_1}}+ \o_1^{r_1+r_2}\s_{\min}(X^{*t})c_0\sqrt{n}\r)^2}.
\end{split}
\end{align}
We note that this upper bound above goes to $\infty$ in the regime $\o_2 \gg \o_1$, which means that gradient descent can avoid lazy training. Note that it does not imply this training scheme is guaranteed to be non-lazy though.

\subsection{Regime with $\o_1 \gg \o_2$ }

In this regime, we have $\ddot{\phi}_{\m}\frac{(\o_2\dot{\phi}_{\m} \s_{\m}(X) + \o_1^{r_2} c_0\sqrt{n})\s_{\m}(X)}{\o_1^{2r_1}\sqrt{d_1}\s_{\min}^2(X^{*t})} \lesssim  \dot{\phi}_{\m}+ \o_2\ddot{\phi}_{\m}\sqrt{d_1}$. Then we have
\begin{align}
\begin{split}
\|h(\Theta_i) - \tilde{h}(\tilde{\Theta}_i)\| & \lesssim \frac{\sqrt{2}\s_{\max}(X)(\dot{\phi}_{\m}+ \o_2\ddot{\phi}_{\m}\sqrt{d_1}) }{(\o_2 \dot{\phi}_{\m}\s_{\m}(X)\sqrt{d_1} + \o_1^{r_2}c_0 \sqrt{(1+\delta_2)d_1n})^2} \\
& \lesssim \frac{\sqrt{2}\s_{\max}(X)(\dot{\phi}_{\m}+ \o_2\ddot{\phi}_{\m}\sqrt{d_1})  }{(\o_1^{r_2}c_0 \sqrt{d_1n})^2}.
\end{split}
\end{align}

Note that this bound goes to 0 and lazy training is bound to happen asymptotically.

\section{Implementation details of Section~\ref{sec:experiment}}
\label{app:exp}
For the experiments illustrated in Figure~\ref{fig:experiment}, we computed the training and test accuracy for different variants of the proposed weight initialization scheme. We considered the MNIST data set made available through the \emph{torchvision} implementation\footnote{This implementation uses the original MNIST source: \url{http://yann.lecun.com/exdb/mnist/}.}. We used the provided split of 60\,000 training examples and 10\,000 test examples which we subsequently normalized.

First, a teacher neural network was train on this data set.
The label provided by the teacher was then used to relabel both the training and test examples.
For each of the weight initializations a student network was constructed and trained on the relabeled data set. The student neural network had 1\,000 units in its hidden layer and used the GeLU activation function.
For the loss we used the mean square error against a one-hot encoding of the true class label.
We minimized this loss with stochastic gradient descent (SGD) for which there was three hyperparameter choices. As the difficult of the data set was modest we expected a large range of these hyperparameters to work. It thus sufficed to make a reasonable guess by choosing a batch size of $128$, learning rate of $0.01$ and $300$ epochs. The teacher neural network differed from the student network by using He initialization and cross entropy loss.

All results were implemented in PyTorch~\citep{torch} and run on a Slurm cluster using a Tesla K40c GPU.
We fixed $\omega_1 \omega_2 \approx 0.002259$ based on the He initialization for our particular network and varied $\o_2$ in the range $[0.002, 0.1]$.
We considered 10 different initialization in this range and ran 5 experiments for each configuration of weight initialization, $(\o_1, \o_2)$.
Using these independent runs we plotted the mean and standard deviation of the final training and test accuracy in Figure~\ref{fig:experiment}, in Section \ref{sec:experiment}.


%

\end{document}